\documentclass[accepted]{uai2025} % after acceptance, for a revised version;
\usepackage[american]{babel}
\usepackage{amsthm,amsmath,amsfonts,amssymb}
\usepackage{natbib} % has a nice set of citation styles and commands
\usepackage{mathtools} % amsmath with fixes and additions
\usepackage{booktabs, tabularx} % commands to create good-looking tables
\usepackage{tikz} % nice language for creating drawings and diagrams
\usepackage{hyperref}
\usepackage{url}
\usepackage{todonotes}
\usepackage{algorithm2e}
\RestyleAlgo{ruled}
\usepackage{amsthm}
\usepackage{bm}
\usepackage{bbm}
\usepackage{float}
\usepackage{tikz}
\usetikzlibrary{arrows}
\usetikzlibrary{shapes}
\usepackage{tkz-berge}

\usetikzlibrary{positioning}
\usetikzlibrary{patterns}
\usetikzlibrary{patterns.meta}
\usetikzlibrary{matrix}
\usetikzlibrary{decorations.pathreplacing}
\usetikzlibrary{graphs,graphs.standard}
\usetikzlibrary{fadings}
\usetikzlibrary{trees}
\usetikzlibrary{mindmap}
\usetikzlibrary{arrows.meta}
\usetikzlibrary{calc}
\usetikzlibrary{fit}
\usetikzlibrary{shapes}

\theoremstyle{plain}

\newtheorem{theorem}{Theorem}[section]
\newtheorem{lemma}[theorem]{Lemma}
\newtheorem{definition}{Definition}
\newtheorem{corollary}{Corollary}
\newtheorem{proposition}{Proposition}
\newtheorem{condition}{Condition}

\theoremstyle{remark}

\newtheorem*{remark}{Remark}

\DeclarePairedDelimiter\abs{\lvert}{\rvert}%
\DeclarePairedDelimiter\norm{\lVert}{\rVert}%

\DeclareMathOperator*{\argmax}{arg\,max}
\DeclareMathOperator*{\argmin}{arg\,min}

\newcommand{\R}{\mathbb{R}}

\newcommand\indep{\protect\mathpalette{\protect\independenT}{\perp}}
\def\independenT#1#2{\mathrel{\rlap{\(#1#2\)}\mkern2mu{#1#2}}}

\title{Nonlinear Causal Discovery for Grouped Data}

\author[1,4]{\href{mailto:<konstantin.goebler@tum.de>?Subject=Your UAI 2025 paper}{Konstantin~Göbler}{}}
\author[3]{Tobias~Windisch}
\author[1,2]{Mathias~Drton}

\affil[1]{%
  Department of Mathematics\\
  TUM School of Computation, Information and Technology\\
  Technical University of Munich
}
\affil[2]{%
  Munich Center for Machine Learning
}
\affil[3]{%
  University of Applied Sciences Kempten\\
  Faculty of Mechanical Engineering
}
\affil[4]{%
  Robert Bosch GmbH
}

\begin{document}
\maketitle

\begin{abstract}
  Inferring cause-effect relationships from observational data has gained significant attention in
recent years, but most methods are limited to scalar random variables. In many important domains, including neuroscience, psychology, social science, and industrial manufacturing, the causal units of interest are groups of variables rather than individual scalar measurements. Motivated by these applications, we extend nonlinear additive noise models to handle random vectors, establishing a two-step approach for causal graph
learning: First, infer the causal order among random vectors. Second, perform model selection to identify the best graph consistent with this order. We introduce effective and novel solutions for both steps in the vector case, demonstrating strong performance in simulations. Finally, we apply our method to real-world assembly line data with partial knowledge of causal ordering among variable groups.

\end{abstract}

\section{Introduction}\label{sec:introduction}

Many techniques have been developed to infer causal relations among random variables from observational data~\citep{Spirtes2000,Chickering2003,Shimizu2006,Peters2014,Zheng2018}. While most of these procedures focus on structure learning among scalar variables, there has been steadily increasing interest in settings where groups of measurements constitute the causal entities of interest~\citep[see][for an overview]{Wahl2024}. For instance in neuroscience, causal connections between brain regions rather than individual neurons are often of interest~\citep{Panzeri2017, Kohn2020, Semedo2020}. In earth and climate science, researchers are frequently interested in causal interactions among groups of measurements (e.g., wind speed, air pressure, etc.) across a number of grid locations spanning the planet \citep[and references therein]{Runge2015}. In psychology and social sciences studies often involve the measurement of certain psychological or societal traits based on several proxy variables that need to be treated jointly \citep{Cronbach1955, Campbell1959, Antonoplis2022}. In industrial manufacturing, quality control systems often record several related measurements from automated process, e.g., industrial welding, injection molding, or staking and pressing, where causal relationships among several such process are the relevant causal items \citep{Vukovic2022,Kikuchi2023,Goebler2024}.

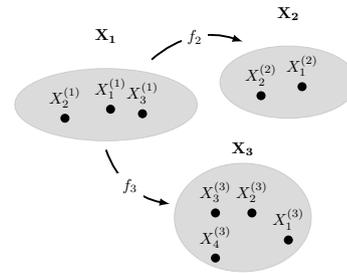
\begin{figure}
  \centering
  \scalebox{0.6}{
    \begin{tikzpicture}[
   ellip/.style = {
       pattern={Lines[distance=5mm,angle=45,line width=0.9mm], }, opacity=0.8
   }
    ]

    % X1
    \draw[fill=gray!60!black, opacity=0.2] (0.0, 0) ellipse (2cm and 0.8cm) coordinate node
    [align=center, text=white, scale=1.0, scale=2] (X1) {};
    \node[above=1cm of X1] {$\mathbf{X_1}$};
    \node[circle, fill=black, inner sep=2pt, label=$X_1^{(1)}$] at (0.1, -0.1) {};
    \node[circle, fill=black, inner sep=2pt, label=$X_2^{(1)}$] at (-0.9, -0.3) {};
    \node[circle, fill=black, inner sep=2pt, label=$X_3^{(1)}$] at (0.8, -0.2) {};

    % X2
    \draw[fill=gray!60!black, opacity=0.2] (4.0, 0.5) ellipse (1.5cm and 0.8cm) coordinate node
    [align=center, text=white, scale=1.0, scale=2] (X2) {};
    \node[above=1cm of X2] {$\mathbf{X_2}$};
    \node[circle, fill=black, inner sep=2pt, label=$X_1^{(2)}$] at (4.3, 0.4) {};
    \node[circle, fill=black, inner sep=2pt, label=$X_2^{(2)}$] at (3.4, 0.2) {};

    % X3
    \draw[fill=gray!60!black, opacity=0.2] (3.0, -2.5) ellipse (1.5cm and 1.2cm) coordinate node
    [align=center, text=white, scale=1.0, scale=2] (X3) {};
    \node[above=1cm of X3] {$\mathbf{X_3}$};
    \node[circle, fill=black, inner sep=2pt, label=$X_1^{(3)}$] at (4.0, -3.0) {};
    \node[circle, fill=black, inner sep=2pt, label=$X_2^{(3)}$] at (3.2, -2.4) {};
    \node[circle, fill=black, inner sep=2pt, label=$X_3^{(3)}$] at (2.4, -2.4) {};
    \node[circle, fill=black, inner sep=2pt, label=$X_4^{(3)}$] at (2.4, -3.4) {};

    \draw[-latex, very thick] (0,-1) to[bend right] node[midway, fill=white] {$f_3$} (1.4, -2.2);
    \draw[-latex, very thick] (1,1) to [bend left] node[midway, fill=white] {$f_2$} (3.0, 1.4);

\end{tikzpicture}
  }
  \caption{An example of a grouped additive noise model with three groups and varying group sizes.}
  \label{fig:ganm_example}
\end{figure}

There are a number of ways to incorporate known group structures in causal learning tasks. The most elementary approach is to apply dimension reduction, e.g., by taking the mean across all members in one group. Given the resulting set of scalar summary variables, standard causal discovery methods may be employed. This comes at the cost of severe information loss, and may even render conditional independence results useless \citep{Wahl2024}. Another approach disregards the grouping structure during the learning tasks and applies causal discovery techniques to the group members. In a second step, the resulting graph estimate is appropriately coarsened to represent the variable groupings~\citep[see][]{Rubenstein2017,Chalupka2016, Parviainen2017, Anand2023}. A last approach seeks to treat the groups themselves as the causal quantities of interest and aims at performing causal learning on the groups directly \citep{Janzing2010, Zscheischler2011, Entner2012, Wahl2023}. Our work belongs to the latter framework of group causal learning.

Motivated by the work of \citet{Peters2014} on nonlinear causal discovery in continuous additive noise models (ANMs) for multivariate data, we revisit the identifiability of causal directions in the group setting. In particular, we show that in general, the causal directions can be identified in the group setting. The statistical problems involved when operating with random vectors rather than scalar random variables become much more challenging. We construct an order independent version of the regression with subsequent independence test (RESIT) algorithm~\citep{Peters2014} in the group setting and propose efficient and flexible solutions to the estimation problems involved.

In particular, the model selection task after having obtained a causal order requires special attention. We propose a novel class of multi-response group sparse additive models (MURGS) for selecting relevant edges. MURGS is a stand-alone feature selection procedure specifically designed for the grouped setting and can be seamlessly integrated into the pruning phase of any order-based method \citep{Teyssier2012}. For example, \citet{Entner2012} estimate causal orderings using grouped linear non-Gaussian ANMs, and we anticipate that other order-based approaches --- such as the score matching method introduced by \citet{Rolland2022} --- can be extended to the group case.

In summary our main contributions are the following:
\begin{itemize}
  \item We propose a flexible and fully nonparametric learning strategy to first obtain a causal order by means of neural networks and nonparametric independence tests. Then, for DAG pruning, we introduce MURGS, a class of multi-response sparse additive models to encourage sparsity on the group level and analytically derive a closed-form backfitting update for the corresponding block coordinate descent algorithm.
  \item We evaluate our proposed method on synthetic data and demonstrate superior performance compared with several other causal discovery algorithms. Further, we consider real world manufacturing data with partially known causal ordering that allows us to partly assess algorithmic performance.
\end{itemize}

The remainder of the paper is structured as follows. Section~\ref{sec:methodology} introduces the group ANM and establishes identifiability results in the group setting. Next, in Section~\ref{sec:groupresit} we describe the two phases of the GroupRESIT algorithm including the development of MURGS for model selection. Section~\ref{sec:experiments} presents the results of our synthetic experiments, and in Section~\ref{sec:real_data} we apply our methods to real data from industrial manufacturing. We conclude in Section~\ref{sec:conclusions}.

\section{Methodology}\label{sec:methodology}

\subsection{Prerequisites}
Throughout, we use the term \emph{group} to refer to a set of scalar random variables that all belong to the same random vector. For integer \(p\) we define \([p] = \{1, \ldots, p\}\). Further, define \(\mathbf{X} = (\mathbf{X}_1, \ldots, \mathbf{X}_p)\) as a tuple of \(p\) random vectors, where each \(\mathbf{X}_g = (X_1^{(g)}, \ldots, X_{d_g}^{(g)})\) is a vector in \(\mathbb{R}^{d_g}\), with \(d_g \in \mathbb{N}\), for \(g \in [p]\). We refer to \(\mathbf{X}_g\) as the \(g\)-th group. Denote by \(P_{\mathbf{X}}\) the joint distribution on \(\mathbb{R}^{d_1} \times \cdots \times \mathbb{R}^{d_p}\). If it exists, we denote the (joint) density of \(\mathbf{X}\) by \(p_{\mathbf{X}}\).
Let \(\mathcal{G} = (V,E)\) be a directed acyclic graph (DAG) with vertex set \(V = [p]\) and edges \(E \subseteq V \times V\).
The vertex set indexes the set of random vectors
in \(\mathbf{X}\), and adopting the framework of graphical models, we assume that \(P_\mathbf{X}\) factorizes according to \(\mathcal{G}\) as follows
\begin{equation*}
  P_{\mathbf{X}}(\mathbf{X}) = \prod_{g=1}^p P_{\mathbf{X}_g}(\mathbf{X}_g \mid \mathbf{X}_{pa_{\mathcal{G}}(g)}),
\end{equation*}
where \(pa_{\mathcal{G}}(g)\) refers to the parents of node \(g\) in \(G\). We refer to \citet{Drton2017} for more details on graphical models.

We call \(j\) a parent of \(g\) if \((j,g) \in
E\). Further, we denote the set of non-descendants by \(nd_{\mathcal{G}}(g) = V\setminus\bigl(\{g\}\cup\mathrm{desc}(g)\bigr)\) where the \emph{descendants} of \(g\) are defined as \(\mathrm{desc}_{\mathcal{G}}(g)=\{w\in V : \text{there exists a directed path from } g \text{ to } w\}\).
Whenever the graph is clear from the context we omit the subscript. Note that the factorization agrees fully with the scalar case as long as \(P_\mathbf{X}\) has a density with respect to some product measure.
We may also express the above model in terms of structural equation models (SEM) in the grouped case:

\begin{definition}\label{def:SEM}
  Let \(\mathbf{X}=(\mathbf{X}_g)_{g\in[p]}\) and
  \(\mathbf{N}=(\mathbf{N}_g)_{g\in[p]}\) be
  jointly distributed random vectors where \(\mathbf{X}_g\) is \(d_g\)-dimensional and where
  \(\mathbf{N}_g \indep \mathbf{N}_h\) for all $g\neq h$.
  If there exists a
  graph \(\mathcal{G}_0\)
  on \([p]\) and a sequence of vector-valued functions \(\mathcal{F}=(f_1,\ldots,f_p)\)
  such that
  \begin{equation}\label{eq:sem}
    \mathbf{X}_g = f_g(\mathbf{X}_{pa_{\mathcal{G}_0}(g)}, \mathbf{N}_g)
  \end{equation}
  for all \(g\in[p]\), then $(\mathbf{X},\mathbf{N}, \mathcal{F},\mathcal{G}_0)$ is a \emph{grouped
  structural equation model} (GSEM).
\end{definition}
In case all groups are one-dimensional, the GSEM is just a standard SEM~\citep{Bollen1989}.
Note that while Definition~\ref{def:SEM}
requires the noise groups to be jointly independent, the model explicitly allows for dependence
within each group.

\subsection{Causal models for grouped data}\label{sec:causal_models}
One can generally not identify the true underlying DAG when given only observational data from the model in Definition~\ref{def:SEM}~\citep[see][for an overview]{Peters2014}. However, as we will see, restricting the functional form of the right-hand side in~\eqref{eq:sem} to be additive in the noise vectors renders the model identifiable.

\begin{definition}\label{def:ANM}
  Let \((\mathbf{X},\mathbf{N}, \mathcal{F},\mathcal{G}_0)\) be a GSEM. If the functions in
  \(\mathcal{F}\) are additive in the noise term, i.e., if
  \[\mathbf{X}_g = f_g(\mathbf{X}_{pa(g)}) + \mathbf{N}_g,\]
  if \(\mathbf{N}_g\) has a strictly positive density with respect to the
  Lebesgue measure for all \(g\in[p]\), and if \(\mathcal{G}_0\) is acyclic,
  then \((\mathbf{X},\mathbf{N}, \mathcal{F},\mathcal{G}_0)\) is a \emph{group
  additive noise model} (GANM).
\end{definition}

In Definition~\ref{def:SEM}, the noise vectors \(\mathbf{N}_g\) and the predictors \(\mathbf{X}_g\) are permitted to have different dimensions, whereas in Definition~\ref{def:ANM} their dimensions must coincide. Moreover, the graph underlying the SEM in Definition~\ref{def:SEM} is not required to be acyclic, while acyclicity is assumed in Definition~\ref{def:ANM}. Figure~\ref{fig:ganm_example} illustrates a GANM with three groups of varying sizes. In addition to the implicit assumption of causal sufficiency—i.e., that no latent variables are present due to the joint independence of the noise terms—the framework for GANMs also requires the notion of causal minimality.

\begin{definition}
  Let \((\mathbf{X}, \mathbf{N}, \mathcal{F}, \mathcal{G}_0)\) be a GANM.
  We say that \(P_\mathbf{X}\) satisfies \emph{causal minimality} if all functions in
  \(\mathcal{F}\) are non-constant in any of their arguments.
\end{definition}

Suppose that the function class \(\mathcal{F}\subseteq C^3(\R^{d_{pa(g)}},\R^{d_g})\), where \(d_{pa(g)}=\sum_{j\in pa(g)}d_j\), consists of nonlinear functions—more precisely, for each output coordinate \(k\in [d_g]\) and each input-dimension \(i\in[d_{pa(g)}]\) there exists some \(\mathbf{x} \in \mathbb{R}^{d_{pa(g)}}\) such that \(\partial^2 f_k/\partial x_i^2(x) \neq 0\) or \(\partial^3 f_k/\partial x_i^3(x) \neq 0\). Then, under causal minimality, identifiability for the bivariate scalar case has been established by~\citet{Hoyer2009}. Consider a
bivariate GANM where the two groups have size~\(1\) respectively, i.e.,
\begin{equation}\label{eq:scalar_bivariate_anm}
  \begin{split}
    X_1 &= N_1 \\
    X_2 &= f_2(X_1) + N_2,
  \end{split}
\end{equation}
where \(X_1 \indep N_2\). Identifiability follows from observing that a regression \(\mathbb{E}[X_2 \mid
X_1] = f_2(X_1)\) along the causal direction leads to independence among the residuals and the predictor
\(X_1\). For general nonlinear functions \(f\), the regression in the anti-causal direction does not lead to independence among the residuals and \(X_2\). Indeed, \citet{Hoyer2009} derive a specific differential equation that the triple \((f_2, P_{X_1}, P_{N_2})\) needs to satisfy for the backwards model to exist. A similar differential equation can be obtained in the bivariate group case.

\begin{condition}\label{cond:identifiability}
  The triple \((f_g, P_{\mathbf{X}_j}, P_{\mathbf{X}_g})\) does not satisfy the following differential equation:
  \begin{multline}\label{eq:tensor_differential_eq}
    D_{\mathbf{x}_j}\mathbf{H}_\xi(\mathbf{x}_j) \Big[(D_{\mathbf{x}_j \mathbf{x}_j}\pi_1(\mathbf{x}_j,\mathbf{x}_g))^{-1} D_{\mathbf{x}_j \mathbf{x}_g}\pi_1(\mathbf{x}_j,\mathbf{x}_g)\Big]
    \\
    \begin{aligned}
      = &D_{\mathbf{x}_j} D_{\mathbf{x}_j \mathbf{x}_g}\pi_1(\mathbf{x}_j,\mathbf{x}_g) - \Big[
        D_{\mathbf{x}_j}\big( \mathbf{H}_{f_g}(\mathbf{x}_j)[\nabla \nu(\mathbf{u})] \big) \\
        - &D_{\mathbf{x}_j}\big( \mathbf{J}_{f_g}(\mathbf{x}_j)^\top \mathbf{H}_{\nu}(\mathbf{u}) \mathbf{J}_{f_g}(\mathbf{x}_j) \big)
      \Big] \\
      \Big[(&D_{\mathbf{x}_j \mathbf{x}_j}\pi_1(\mathbf{x}_j,\mathbf{x}_g))^{-1} D_{\mathbf{x}_j \mathbf{x}_g}\pi_1(\mathbf{x}_j,\mathbf{x}_g)\Big],
    \end{aligned}
  \end{multline}
  where \(\mathbf{u} \coloneqq \mathbf{x}_g-f_g(\mathbf{x}_j)\) and further \(\nu \coloneqq \log p_{\mathbf{N}_g}\), \(\xi \coloneqq \log p_{\mathbf{X}_j}\), with arguments \(\mathbf{u}, \mathbf{x}_j\), respectively. Additionally, \(\pi_1({\mathbf{x}_j,\mathbf{x}_g}) \coloneqq \log p_{\mathbf{X}_j,\mathbf{X}_g}(\mathbf{x}_j, \mathbf{x}_g)\) and
  \begin{align*}
    D_{\mathbf{x}_j \mathbf{x}_j}\pi_1(\mathbf{x}_j,\mathbf{x}_g) &= \mathbf{H}_{\xi}(\mathbf{x}_j) - \mathbf{H}_{f_g}(\mathbf{x}_j)[\nabla \nu\bigl(\mathbf{u}\bigr)] \\
    &+ \mathbf{J}_{f_g}(\mathbf{x}_j)^\top \mathbf{H}_{\nu}\bigl(\mathbf{u}\bigr) \mathbf{J}_{f_g}(\mathbf{x}_j),
  \end{align*}
  and \(D_{\mathbf{x}_j \mathbf{x}_g}\pi_1(\mathbf{x}_j,\mathbf{x}_g) = -\mathbf{J}_{f_g}(\mathbf{x}_j)^\top \mathbf{H}_{\nu}\bigl(\mathbf{u}\bigr)\). \(\mathbf{J}\) and \(\mathbf{H}\) denote Jacobian and Hessian matrices or tensors, respectively.
\end{condition}

\begin{remark}
  Note that when \(\mathbf{J}_{f_g}(\mathbf{x}_j)\) is full rank and \(\mathbf{H}_{\nu}(\mathbf{u})\) is positive definite, we can fully isolate the third-order derivative tensor \(D_{\mathbf{x}_j}\mathbf{H}_\xi(\mathbf{x}_j)\):
  \begin{equation*}
    \begin{split}
      D_{\mathbf{x}_j}\mathbf{H}_\xi(\mathbf{x}_j) &= D_{\mathbf{x}_j}(D_{\mathbf{x}_j\mathbf{x}_g}\pi_1)(D_{\mathbf{x}_j\mathbf{x}_g}\pi_1)^{-1}D_{\mathbf{x}_j\mathbf{x}_j}\pi_1 \\
      &- \Big[D_{\mathbf{x}_j}(\mathbf{H}_{f_g}(\mathbf{x}_j)[\nabla \nu(\mathbf{u})]) \\
      &- D_{\mathbf{x}_j}(\mathbf{J}_{f_g}(\mathbf{x}_j)^\top \mathbf{H}_{\nu}(\mathbf{u}) \mathbf{J}_{f_g}(\mathbf{x}_j))\Big],
    \end{split}
  \end{equation*}
  where we have suppressed the arguments $\mathbf{x}_j$ and $\mathbf{x}_g$ of \(\pi_1\). Observe
  that \(\mathbf{H}_\xi(\mathbf{x}_j)\) enters only via the term
  \(D_{\mathbf{x}_j\mathbf{x}_j}\pi_1\) on the right hand side. In fact, this equation bears a direct resemblance to the scalar form derived by~\citet{Hoyer2009}, highlighting that our result is a natural generalization of the scalar case. In general, Eq.~\eqref{eq:tensor_differential_eq} describes a directional projection of \(D_{\mathbf{x}_j}\mathbf{H}_\xi(\mathbf{x}_j)\) onto the directions defined by the columns of the matrix \((D_{\mathbf{x}_j\mathbf{x}_j}\pi_1)^{-1} D_{\mathbf{x}_j\mathbf{x}_g}\pi_1\). The dimensions \(d_{x_j}\) and \(d_{x_g}\) determine the range of the resulting tensor contraction.
  A detailed exploration of the role of the group sizes and implications for the form of the triple \((f_g, P_{\mathbf{X}_j}, P_{\mathbf{X}_g})\) is deferred to future work.
\end{remark}

\begin{definition}
  Consider a \emph{bivariate GANM} given by the equations
  \begin{equation*}
    \mathbf{X}_j = \mathbf{N}_j,\quad\mathbf{X}_g = f_g(\mathbf{X}_j) + \mathbf{N}_g,
  \end{equation*}
  for \(\{j,g\} = \{1,2\}\). If the corresponding triple \((f_g, P_{\mathbf{X}_j}, P_{\mathbf{X}_g})\) satisfies Condition~\ref{cond:identifiability}, we call this model an \emph{identifiable bivariate} GANM.
\end{definition}

\begin{theorem}\label{theorem:bivariate_identifiability}
  Let \(P_{\mathbf{X}}\) be the joint distribution of \(\mathbf{X}\) generated by an
  \emph{identifiable bivariate} GANM \((\mathbf{X}, \mathbf{N}, \mathcal{F}, \mathcal{G}_0)\)
  and suppose that causal minimality holds. Then, the graph \(\mathcal{G}_0\) is
  identifiable from \(P_{\mathbf{X}}\).
\end{theorem}

All proofs for this section are provided in Appendix~\ref{sec:proofs_A}. As demonstrated by \citet{Peters2014}, extending the analysis from two to multiple variables can be achieved by appropriately constraining the involved distributions and functions so that, locally, the problem reduces to the bivariate case.
\begin{corollary}\label{corollary:multivariate_identifiability}
  Let \((\mathbf{X}, \mathbf{N}, \mathcal{F}, \mathcal{G}_0)\) be a GANM with \(p\) variables. Suppose that for each node \(g \in [p]\), each parent \(j \in pa(g)\), and every set \(S\) satisfying
  \begin{equation*}
    pa(g) \setminus \{j\} \subseteq S \subseteq nd(g) \setminus \{g,j\},
  \end{equation*}
  there exists a realization \(\mathbf{x}_S \in \mathbb{R}^{\abs{S}}\) for which the conditional distribution of \(\mathbf{X}_S\) has strictly positive density with respect to the Lebesgue measure, and the triple
  \begin{equation*}
    \left(f_g(\mathbf{x}_{pa(g)\setminus \{j\}}, \mathbf{x}_j), P_{\mathbf{X}_j,\mathbf{X}_g\mid \mathbf{X}_S=\mathbf{x}_S}, P_{\mathbf{X}_j}\right)
  \end{equation*}
  satisfies Condition~\ref{cond:identifiability}. Here, the arguments \(\mathbf{x}_{pa(g)\setminus \{j\}}\) are held fixed, making the function \(f_g(\mathbf{x}_{pa(g)\setminus \{j\}}, \mathbf{x}_j)\) depend solely on \(\mathbf{x}_j\).

  Then the graph \(\mathcal{G}_0\) is identifiable from the joint distribution \(P_{\mathbf{X}}\).
\end{corollary}

Clearly, fixing all arguments of \(f_g\) except \(\mathbf{X}_j\) also leads to a bivariate GANM. Corollary~\ref{corollary:multivariate_identifiability} suggests that this is not enough. Instead, we need to put restrictions on the conditional distribution of \(\mathbf{X}_j\). The following result guides the estimation strategy described in the ensuing section.

\begin{lemma}\label{lemma:ancestor_independence}
  Let $(\mathbf{X}, \mathbf{N}, \mathcal{F}, \mathcal{G}_0)$ be a GSEM. For all \(S \subseteq
  nd_{\mathcal{G}_0}(g)\), we have \(\mathbf{N}_g \indep \mathbf{X}_S\).
\end{lemma}

In particular, if \(g \in [p]\) is a sink node, i.e., a node without any descendants, its
non-descendants is the set of all other nodes in \(\mathcal{G}_0\). Thus, its corresponding noise
vector \(\mathbf{N}_g\) is independent of all other variables \(\mathbf{X}_{[p]\setminus
\{g\}}\).

\section{Grouped RESIT}\label{sec:groupresit}

In the context of ANMs, \citet{Peters2014} introduce the Regression with Subsequent Independence
Test (RESIT) algorithm to learn the structure of the underlying DAG from observational data. In what follows, we denote the underlying true DAG by \(\mathcal{G}_0\). RESIT
has two phases. The \textbf{first phase} infers a causal order among the variables
involved. A permutation \(\pi : [p] \to [p]\) is a valid causal order for a DAG \(\mathcal{G}\)
if \(\pi(g) < \pi(j)\) for all \(g\in nd_{\mathcal{G}}(j)\). Motivated by
Lemma~\ref{lemma:ancestor_independence}, the following steps are performed \(p-1\) times:
\begin{enumerate}
  \item[(i)] Regress each variable on all others. Then measure independence between the corresponding residuals and all other variables.
  \item[(ii)] The variable that accounts for its \textit{least dependent residual} is considered as a sink node and removed from the set of variables.
\end{enumerate}

The \textbf{second phase} seeks to select the \textit{best} DAG among those that agree with the causal order found in the first phase. Several model selection techniques may be used for this purpose.~\citet{Peters2014} propose to \textit{greedily test away} edges that are present in the super-DAG associated with the inferred causal order. The strategy we pursue involves sparse model selection techniques to prune extraneous edges.

In our setting, we can apply the same two-phase procedure to obtain a group DAG estimate of \(\mathcal{G}_0\). Yet, all involved estimation and testing procedures need to solve much harder statistical problems. We propose solutions to the estimation tasks in \textsc{Phase I} and \textsc{II} (Algorithms~\ref{alg:gRESIT_phase1} and~\ref{alg:gRESIT_phase2}).

\subsection{Learning a Causal Order}

\begin{algorithm}
  \caption{GroupRESIT algorithm \textsc{Phase I}}
  \SetKwInOut{Input}{Input}
  \SetKwInOut{Output}{Output}
  \SetKwInOut{Initialize}{Initialize}
  \Input{IID samples of p-many jointly distributed random vectors \((\mathbf{X}_1, \ldots, \mathbf{X}_p)\).}
  \Output{Learned causal order \(\pi\).}

  \Initialize{\(S \coloneq \{1, \ldots, p\}, \quad  \pi \coloneq [\cdot]\).}
  \Repeat{until \(S = \emptyset\)}{
    \For{\(g\in S\)}{
      \begin{itemize}
        \item[(i)] Regress \(\mathbf{X}_g = (X_1^{(g)}, \ldots, X_{d_g}^{(g)})\) \\
          on \(\{\mathbf{X}_j\}_{j\in S \setminus \{g\}}\)
        \item[(ii)] Measure the independence between \\
          the residuals \(\mathbf{R}_g = (R_1^{(g)}, \ldots, R_{d_g}^{(g)})\) \\
          and the remaining groups \(\{\mathbf{X}_j\}_{j\in S \setminus \{g\}}\)
      \end{itemize}
    }
    Identify the group \(g^*\) that accounts for its least dependent residual\\
    \(S \coloneq S \setminus \{g^*\}\) \\
    \(\pi \coloneq [g^*, \pi]\)
  }
  \label{alg:gRESIT_phase1}
\end{algorithm}

The first phase (Algorithm~\ref{alg:gRESIT_phase1}) consists of repeatedly solving regression
problems on progressively smaller predictor sets. In each iteration, we train multi-output deep
neural networks \(\phi_\theta\) whose output layers are sized to match the number of response variables
in each group.

After training is completed, we compute the residuals
\(\mathbf{R}_g = \phi_\theta(\mathbf{X}_g) - \mathbf{X}_g,\)
and identify the group whose residuals exhibit the weakest dependence on the remaining variables. This group is then designated as a \emph{sink node} and is subsequently removed from the set of variables.

We measure dependence via the HSIC~\citep{Gretton2005},
i.e.,
\(\text{HSIC}((\mathbf{X}_j)_{{j\in S \setminus \{g\}}}, \mathbf{R}_g)\).
When equipped with characteristic kernels \citep{Fukumizu2007}, the HSIC equals
zero if and only if the random quantities involved are (unconditionally) independent. In particular, the HSIC is applicable to random vectors of general dimension.

\begin{remark}
  \noindent We emphasize that direct comparisons of HSIC values are meaningful only when all models under consideration share the same dimension and all involved quantities are measured on an identical scale.
\end{remark}

\subsection{MURGS for model selection}\label{sec:murgs}

Having obtained a causal order \(\pi\), we construct a super-DAG \(\mathcal{G}^{\pi}\) where nodes get assigned all of their predecessors in the causal order as parents.
Formally, the set of parents for node \(j\in [p]\) in \(\mathcal{G}^{\pi}\) is \(pa_{\pi}(j) = \{g \in [p]: \pi(g) < \pi(j)\}\). The goal of \textsc{Phase II} is to find a not too large super-DAG \(\mathcal{G}\) of \(\mathcal{G}_0\) with \(\mathcal{G} \subseteq \mathcal{G}^{{\pi}}\). We advocate to use sparse model selection techniques for this procedure. To that end, we tailor multi-task sparse additive models to the given setting. The resulting model class is of independent interest as it generalizes the sparse group lasso to the multi-task setting. In this paper, we use a common design matrix across all tasks rendering the models multi-response in nature.

In the original RESIT, \citet{Peters2014} propose to iteratively remove nodes from the potential parent set by greedily cycling through the following steps. First, remove a potential parent node from the regression set and second, test whether residuals are still independent and restore the node in question if this is not the case. Thus, the significance level of the test acts as a tuning parameter for the model selection procedure. As pointed out in \citet{Peters2014}, such a procedure strongly depends on the order in which the independence tests are carried out. Accordingly, type-one errors lead to extraneous edges in the final DAG estimate.
Instead, we use feature selection via sparse additive models to prune edges in
\({\mathcal{G}}^{\pi}\).

\begin{algorithm}
  \caption{GroupRESIT algorithm \textsc{Phase II}}
  \SetKwInOut{Input}{Input}
  \SetKwInOut{Output}{Output}
  \SetKwInOut{Initialize}{Initialize}
  \Input{\(\pi\)}
  \Output{Learned DAG \({\mathcal{G}}\)}
  \Initialize{\(pa_\pi\).}
  \For{\(j \in \pi\)}{
    \begin{itemize}
      \item Use MURGS and regress \(\mathbf{X}_j\) on \(\{\mathbf{X}_g\}_{g \in pa_\pi(j)}\)
      \item Obtain the parent set \(pa_{\mathcal{G}}(j) \coloneqq \{g\in pa_\pi(j): f^{(k)}_{j,g,h} \neq 0\}\)
    \end{itemize}
  }
  \label{alg:gRESIT_phase2}
\end{algorithm}

More explicitly, we are interested in the best sparse approximation
of the regression function \(\mathbb{E}[\mathbf{X}_{j} \mid \mathbf{X}_{pa_{\mathcal{G}^{\pi}}(j)} = \mathbf{x}]\) of the form
\begin{equation*}
  \mathbb{E}[{X}_{k}^{(j)} \mid \mathbf{X}_{pa_{\mathcal{G}^{\pi}}(j)} = \mathbf{x}] \approx \sum_{g \in pa_{\mathcal{G}^{\pi}}(j)} \sum_{h \in [d_g]} f_{j,g,h}^{(k)}(x^{(g)}_h),
\end{equation*}
for \(k \in [d_j]\). The first index \(j\) identifies the node whose incoming edges are subject to pruning. The second index \(g\) corresponds to the parent groups, and the third index \(h\) to the individual entries within each parent group. Finally, the superscript \((k)\) selects the respective entry within the response group. To facilitate feature selection, we introduce a regularization functional that encourages a common sparsity pattern shared by both predictor and response group elements.

\begin{figure}
  \centering
  \begin{tikzpicture}
  \node[] (xg) at (0, -0.2) {$\mathbf{X_g}$};

  \node[] (g1) at (0, -1) {$X^{(g)}_{1}$};
  \node[scale=0.5]  at (0, -1.5) {$\vdots$};
  \node[] (gh) at (0, -2) {$X^{(g)}_{h}$};
  \node[scale=0.5]  at (0, -2.5) {$\vdots$};
  \node[] (gd) at (0, -3.0) {$X^{(g)}_{d_g}$};

  \node[] (xj) at (3, -0.2) {$\mathbf{Y}$};
  \node[] (j1) at (3, -1) {$Y^{(1)}$};
  \node[scale=0.5]  at (3, -1.5) {$\vdots$};
  \node[] (jk) at (3, -2) {$Y^{(k)}$};
  \node[scale=0.5]  at (3, -2.5) {$\vdots$};
  \node[] (jd) at (3, -3.0) {$Y^{(d_j)}$};

  \draw[-latex, very thick] (xg) to (xj);

  \node[]  at (1.5, -1.1) {$\mathbf{f^{(k)}_g}$};

  \draw[-latex] (g1) to (jk);
  \draw[-latex] (gh) to node[midway, fill=white] {$f_{g,h}^{(k)}$} (jk);
  \draw[-latex] (gd) to (jk);

\end{tikzpicture}
  \caption{Overview of MURGS notation.}
  \label{fig:notation}
\end{figure}
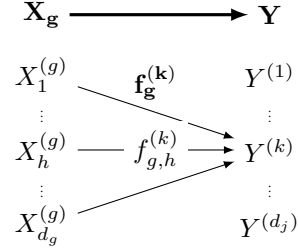

Provided we have found such a functional we remove spurious edges among possible parents \(g \in pa_{\mathcal{G}^{\pi}}(j)\) to obtain a set of relevant edges
\begin{equation*}
  {E}^{\pi} = \{(g,j): f_{j,g,h}^{(k)} \neq 0, \forall k \in [d_j] \text{ and } h\in[d_g]\}.
\end{equation*}
Algorithm~\ref{alg:gRESIT_phase2} summarizes the procedure.

\subsection{Simultaneous Sparse Backfitting}

Before stating the problem more formally, we introduce some relevant notation.
If some random variable \(Z\) has a distribution \(P_Z\), and \(f\) is a function of \(z\), we
denote its \(L_2(P_Z)\) norm by \(\norm{f}^2:= \int_\mathcal{Z} f^2(z) dP_Z = \mathbb{E}[f^2]\).
For an \(n\)-dimensional vector \(\mathbf{\nu} = (\nu_1,
\ldots, \nu_n)\in\mathbb{R}^n\) we define \(\norm{\mathbf{\nu}}_2^2 = \frac{1}{n} \sum_{i=1}^n \nu_i^2\) and
\(\norm{\mathbf{\nu}}_\infty = \max_{i\in[n]} \abs{\nu_i}\). Consider the \(p\)-dimensional random vector
\(\mathbf{Z} = (Z_1, \ldots, Z_p)^T\). Denote by \(\mathcal{H}_i, i\in[p]\) the Hilbert subspace
\(L_2(P_{Z_i})\) of \(P_{Z_i}\)- measurable functions \(f_i(z_i)\) of the scalar variable \(Z_i\)
with \(\mathbb{E}[f_i(z_i)] = 0\). Hence, \(\mathcal{H}_i\) is equipped with the inner product
\(\langle f_i,g_i \rangle = \mathbb{E}[f_i(Z_i)\cdot g_i(Z_i)]\). Whenever a quantity is estimated from a finite sample of size \(n\), we denote this estimate with a hat.

Furthermore, to ease notation in what follows, we take a closer look at node \(j \in [p]\) and its parents \(pa_j \coloneqq pa_{\mathcal{G}^{\pi}}(j)\). We define \(Y^{(k)} \coloneqq X_{k}^{(j)}\) such that we may drop the corresponding subscripts, i.e. \(f_{g,h}^{(k)} \coloneqq f_{j,g,h}^{(k)}\). Further, we define \(f^{(k)}(x) = \sum_{g \in pa_j}\sum_{h \in [d_g]}
f_{g,h}^{(k)}(x^{(g)}_h)\) for \(k\in [d_j]\). Denote by \(\mathcal{L}_{f^{(k)}}(x,y) = (y^{(k)} - f^{(k)}(x))^2\) the quadratic loss.
We often write \(f_{g,h}^{(k)} \coloneq f_{g,h}^{(k)}(X^{(g)}_h)\). For group \(g \in pa_j\), we denote by \(\mathbf{f}_g^{(k)} = (f_{g,1}^{(k)}, \ldots, f_{g,d_g}^{(k)})^T\) the vector of group component functions and let \(\norm{\mathbf{f}_g^{(k)}} = \sqrt{\sum_{h=1}^{d_g} \norm{f_{g,h}^{(k)}}^2 }\). Figure~\ref{fig:notation} visualizes the notation.

The main objective is to perform feature selection among groups of variables. We formulate a regularization scheme to encourage joint functional sparsity. However, the component functions are allowed to vary among response group members and predictor groups while sharing a common sparsity pattern. To achieve this, consider the following regularization functional
\begin{equation*}
  \Phi^j(f) = \sum_{g \in pa_j} \sqrt{d_g} \max_{k \in [d_j]} \norm{\mathbf{f}_g^{(k)}}.
\end{equation*}
The functional \(\Phi^j(f)\) combines the sum of sup-norms regularization with the
functional version of the \(\ell 1 / \ell 2\) norms. Similar to the group lasso
\citep{Yuan2006}, the \(\ell 1 / \ell 2\) norm induces sparsity at the group level. In turn, the
sup-norm penalty encourages sparsity among the \(d_j\) response group components. A group of
component functions across \(k \in [d_j]\) is removed if and only if all involved smooth functions are
estimated to be zero. On the other hand, if a component function group \(g\) is important with a
positive sup-norm for some response \(k\), no additional penalty is imposed on the \(d_j-1\) remaining
ones. For feature selection purposes, this is desirable as all function groups remain in the model
as long as the sup-norm is positive. This is in contrast to imposing the sum of \(\ell_2\) norms
across the \(d_j\) responses which is often applied in multi-task settings \citep[see
e.g.][]{Argyriou2006, Liu2009, Wang2011, Li2020}.

If the response is a scalar random variable, MURGS reduces to the group sparse additive model treated in~\citet{Yin2012}. In turn, if all groups \(g \in pa_j\) contain only singletons this model reduces to the sparse additive model proposed by~\citep{Liu2007,Ravikumar2009}. If only the response variable is vector-valued and the remaining groups are singletons this model reduces to the multi-response sparse additive model discussed by~\citet{Liu2008}.

MURGS can be cast as a penalized \(\mathbf{M}\)-estimator~\citep{Negahban2012} through the following optimization problem
\begin{equation*}\label{eq:m_estimator}
  \hat{\mathbf{f}} = \min_{\mathbf{f} : f_{g,h}^{(k)} \in \mathcal{H}_{g,h}}
  \Bigg\{ \frac{1}{2n} \sum_{\substack{k\in[d_j], \\ i\in[n]}}\mathcal{L}_{f^{(k)}}(\mathbf{x}_i, y_i^{(k)})
  + \lambda \Phi^j(f) \Bigg\}
\end{equation*}
with \(\lambda > 0\) a regularization parameter.

\begin{algorithm}
  \caption{SoftThresholding update for group \(j\)}\label{alg:soft_thresholding}
  \SetKwInOut{Input}{Input}
  \SetKwInOut{Output}{Output}
  \Input{Partial residual \(\hat{R}_g^{(k)}\) for \(k \in [d_j]\), smoother matrices \(\{S_{g,h} : h\in [d_g]\}\), and tuning parameter \(\lambda\).}
  \Output{\(\mathbf{\hat{f}}_g^{(k)}\) = \((\hat{f}_{g,h}^{(k)})_{h \in [d_g]}\) for \(k \in [d_j]\).}
  Estimate \(P_h R_g^{(k)}\) by smoothing: \(\hat{P}_h^{(k)} = S_{g,h}\hat{R}_g^{(k)}\). \\
  Estimate \(s_g^{(k)} = \norm{\mathbf{Q}R_g^{(k)}}\) by:
  \(\hat{s}_g^{(k)} = \Big({1}/{n} \sum_{h \in [d_g]} \norm{\hat{P}_h^{(k)}}^2\Big)^{1/2} \). \\
  \eIf{\(\sum_{k\in K} s_g^{(k)} \leq \sqrt{d_g}\lambda \)}
  {Set \(\mathbf{\hat{f}}_g^{(k)} = 0\) for all \(k \in [d_j]\).}{Order the indices according to
    \(\hat{s}_g^{(k_1)} \geq \hat{s}_g^{(k_2)} \geq \cdots \geq \hat{s}_g^{(k_{d_j})}\). \\
    Set \(m^* = \argmax_{m} \frac{1}{m} \left( \sum_{l = 1}^{m^*} \hat{s}_g^{(k_{l})} - \sqrt{d_g}\lambda \right)\) \\
    \begin{equation*}
      \hat{f}_{g,h}^{(k_i)} =
      \begin{cases}
        \hat{P}_h^{(k_i)}                                                                                                                   \qquad \text{for } i > m^*&    \\
        \begin{aligned}
          \frac{1}{m^*}\Bigg[ &\sum_{l = 1}^{m^*} \hat{s}_g^{(k_{l})}
          - \sqrt{d_g}\lambda \Bigg] \frac{\hat{P}_h^{(k_i)}}{\hat{s}_g^{(k_{i})}}
        \end{aligned}
        &\text{o.w.}
      \end{cases}
    \end{equation*}
  }
  Center \(\hat{f}_{g,h}^{(k)}\) by subtracting its mean.
\end{algorithm}
\paragraph{Block-Coordinate Descent Algorithm}

In order to solve the optimization problem above, we employ a block-coordinate descent algorithm~\citep[see e.g.][]{Hastie2015}.
First, we derive the population version of the estimation problem. This leads to a range of sub-problems in each iteration that can be solved by means of a soft-thresholding update. Similar solutions in multi-task scalar settings have been found in the linear case as well as for sparse additive models~\citep{Liu2009a,Liu2008}. As is common in backfitting algorithms, we obtain a finite sample version of the algorithm by replacing the conditional expectations with nonparametric smoothers.

Consider the partial residual \(R_g^{(k)} = Y^{(k)} - \sum_{g' \neq g} \sum_{h \in [d_{g'}]} f_{g',h}^{(k)}\) and assume that functions in group \(g\) can be fixed. Then the optimization problem on the population level cuts down to
\begin{multline}\label{eq:backfitting}
  \mathbf{f}_g =
  \argmin_{\mathbf{f}_g : f_{g,h}^{(k)} \in \mathcal{H}_{g,h}} \Bigg\{ \frac{1}{2}
    \mathbb{E}\Big[ \sum_{k=1}^{d_j} \big(R_g^{(k)} - \sum_{h\in [d_g]} f_{g,h}^{(k)}\big)^2 \Big] \\
  + \lambda \sqrt{d_g} \max_{k \in [d_j]} \norm{\mathbf{f}_g^{(k)}}\Bigg\}.
\end{multline}
Now, we are ready to state the population block update.
\begin{theorem}\label{thm:backfitting_update}
  Denote \(P_h = \mathbb{E}[\ \cdot \mid X_h^{(g)}]\) the conditional expectation operator,
  \(\mathbf{Q} = (P_h)_{h \in [d_g]}\) and \(s_g^{(k)} = \norm{\mathbf{Q}R_g^{(k)}}\). Assume that
  \(\mathbb{E}[f_{g,h'}^{(k)} \mid X_{h}^{(g)}] =  0\) for all \(h' \neq h\), i.e., the covariance
  among the component functions within groups is zero. Order the indices according to \(s_g^{(k_1)}
  \geq s_g^{(k_2)} \geq \cdots \geq s_g^{(k_{d_j})}\). Then the solution to Eq.~\eqref{eq:backfitting} has coordinate functions given by
  \begin{equation*}
    f_{g,h}^{(k_i)} = P_h^{(k_i)}R_g^{(k_i)}
  \end{equation*}
  if \(i > m^*\) and by
  \begin{equation*}
    f_{g,h}^{(k_i)} = \frac{1}{m^*}\Bigg[ \sum_{l=1}^{m^*} s_g^{(k_{l})} - \sqrt{d_g}\lambda \Bigg]_{+} \frac{P_h^{(k_i)}R_g^{(k_i)}}{s_g^{(k_i)}}
  \end{equation*}
  if \(i \leq m^*\). Here, \(h \in [d_g]\) and
  \begin{equation*}
    m^* = \argmax_{m \in [d_j]} \frac{1}{m} \left( \sum_{l=1}^m s_g^{(k_{l})} - \sqrt{d_g}\lambda \right),
  \end{equation*}
  with \([\,\cdot\,]_+\) denoting the positive part function.
\end{theorem}

The proof involves calculus of variations in Hilbert spaces and is given in Appendix~\ref{sec:backfitting_update}.
The zero covariance assumption is crucial to obtain a closed form update~\citep[see][]{Foygel2010}. However, for feature
selection we are primarily interested in the case where the sup-norm subdifferential evaluated at
\((\norm{\mathbf{f}_g^{(1)}}, \ldots, \norm{\mathbf{f}_g^{(d_j)}})^T\) is the zero vector. It turns
out that the condition for this case holds in general without any assumptions on the conditional
expectation. The following result makes this explicit:
\begin{proposition}\label{prop:all_zeros}
  \(\norm{\mathbf{f}_g^{(k)}} = 0\) for all \(k \in [d_j]\) if and only if \(\sum_{k=1}^{d_j} \norm{\mathbf{Q}R_g^{(k)}} \leq \lambda \sqrt{d_g}\).
\end{proposition}
Once the stationary condition is derived, the proof to the proposition is straight-forward and can also be found in Appendix~\ref{sec:backfitting_update}.
Based on Theorem~\ref{thm:backfitting_update}, Algorithms~\ref{alg:soft_thresholding} and~\ref{alg:backfitting} detail the backfitting algorithm for MURGS in the finite sample setting.
\begin{algorithm}
  \caption{Backfitting algorithm}\label{alg:backfitting}
  \SetKwInOut{Input}{Input}
  \SetKwInOut{Output}{Output}
  \SetKwInOut{Initialize}{Initialize}
  \Input{Data \(\mathbf{X} = \{\mathbf{X}_g \in \mathbb{R}^{n\times d_g}: g \in pa_j\}\), \(Y \in \R^{n \times d_j}\), regularization parameter \(\lambda\).}
  \Output{Fitted functions \(\hat{\mathbf{f}} = (\hat f_{g,h}^{(k)})_{g \in pa_j, h \in [d_g], k \in
  [d_j]}\).}
  \Initialize{\(\hat{\mathbf{f}} = \mathbf{0}\) pre-compute the smoother matrices \(\{S_{g,h} \in \mathbb{R}^{n\times n} : h \in {d_g}, g\in pa_j\}\).}
  % \While{\(t \leq \text{max\_iter}  \ \& \  \text{incr} > \text{tol}\)}{
  \For{\(g \in pa_j\) until convergence}{

    \begin{itemize}
      \item[(i)] Update partial residual \(\hat{R}_g^{(k)} = Y^{(k)} - \sum_{g' \neq g} \sum_{h \in [d_{g'}]} \hat{f}_{g',h}^{(k)}\)
      \item[(ii)] \(\mathbf{\hat{f}}_g^{(k)} \gets \text{SoftThresholding}(\hat{R}_g^{(k)}, S_{g,:}, \lambda\))
    \end{itemize}

  }
\end{algorithm}
We choose the regularization parameter \(\lambda\) by the generalized cross validation (GCV) criterion from \citet{Liu2007} and adapted to the multi-response setting by \citet{Liu2008}. In the multi-response case, the GCV criterion is given by
\begin{equation*}
  \text{GCV}(\lambda) = \frac{1}{n} \sum_{i=1}^n \frac{\sum_{k=1}^{d_j}
  \mathcal{L}_{\hat{f}^{(k)}}(\mathbf{x}_i, y_i^{(k)})}{(n^2d_j^2 - (nd_j)\text{df}(\lambda))^2},
\end{equation*}
where \(\text{df}(\lambda) = d_j\sum_{g\in d_i} \nu_g I(\sum_{k=1}^{d_j}\norm{
\mathbf{f}_{g}^{(k)}} \neq 0)\) and \(\nu_g = \sum_{h \in [d_g]} \text{tr}(S_{g,h})\) denotes the effective degrees of freedom for the local linear smoother \(S_{g,h}\).

\section{Experiments}\label{sec:experiments}

\paragraph{Setup}
\begin{figure}
  \includegraphics[width=.48\textwidth]{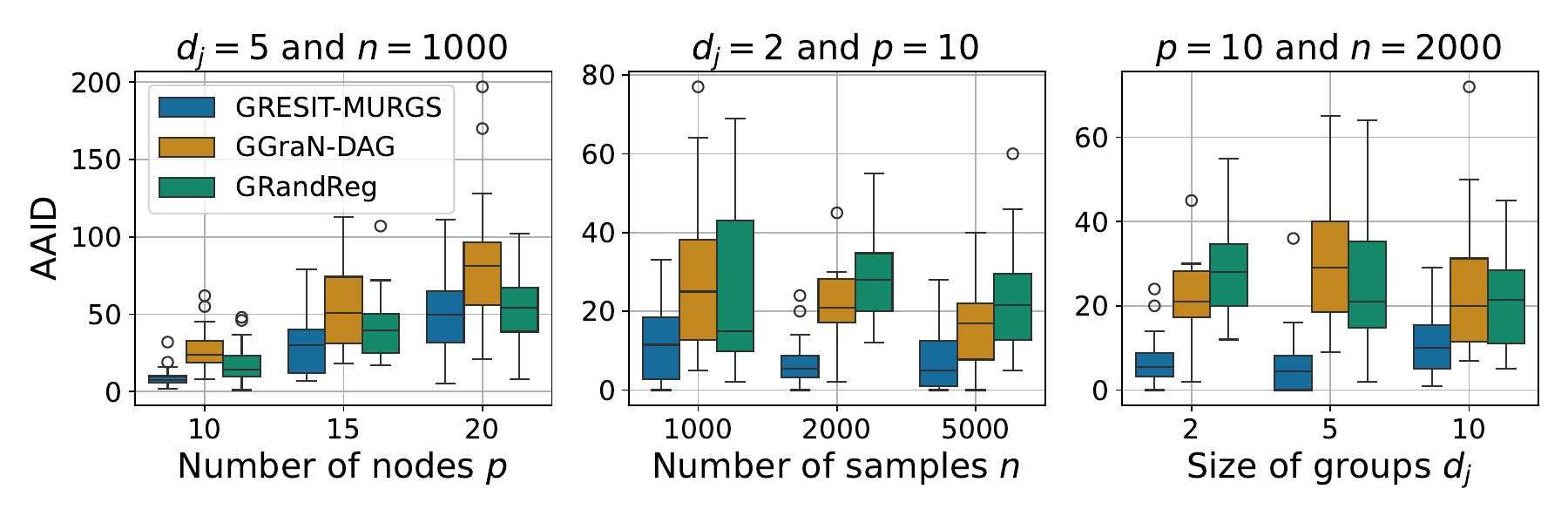}
  \caption{Performance of the algorithms under varying dimensions of the GANM.}
  \label{fig:benchmark_scaling}
\end{figure}

In this section we assess and compare the performance of GroupRESIT with MURGS (\textit{GRESIT-MURGS}), GroupRESIT with greedy independence testing (\textit{GRESIT-IND}), a grouped version of the PC algorithm~\citep{Spirtes1993} \textit({GPC}) and a grouped version of GraN-DAG~\citep{Lachapelle2020} (\textit{GGraN-DAG}). Furthermore, we report a baseline algorithm that picks a causal order at random and subsequently applies MURGS (\textit{GRandReg}). Implementation details—including our modifications to the PC algorithm and GraN-DAG to accommodate the group setting, along with a description of the metrics, synthetic data, and a table of hyperparameters—are provided in Section~\ref{sec:simulation_details}.

\begin{figure*}
  \centering
  \includegraphics[width=.82\textwidth]{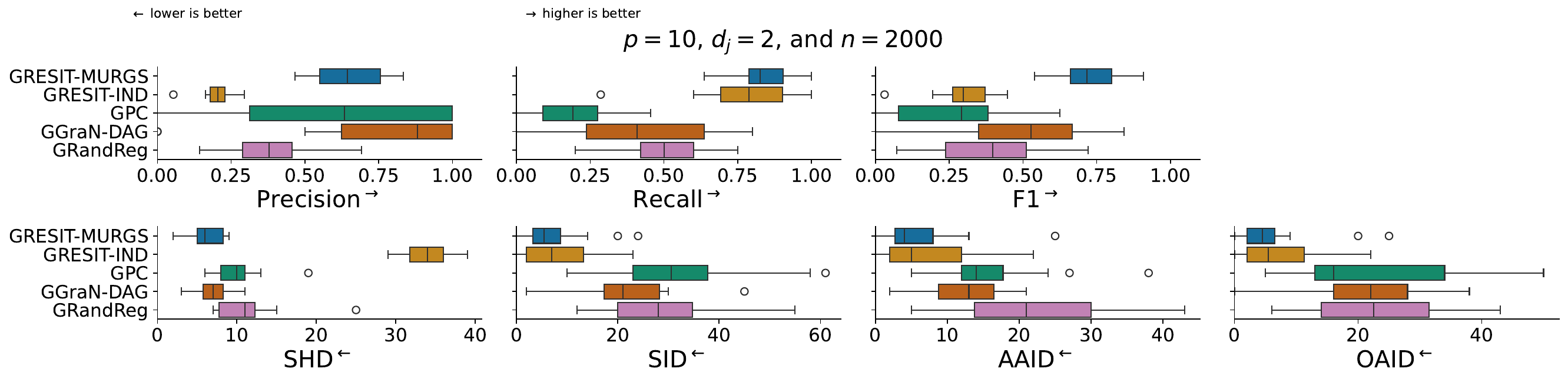}\\
  \includegraphics[width=.82\textwidth]{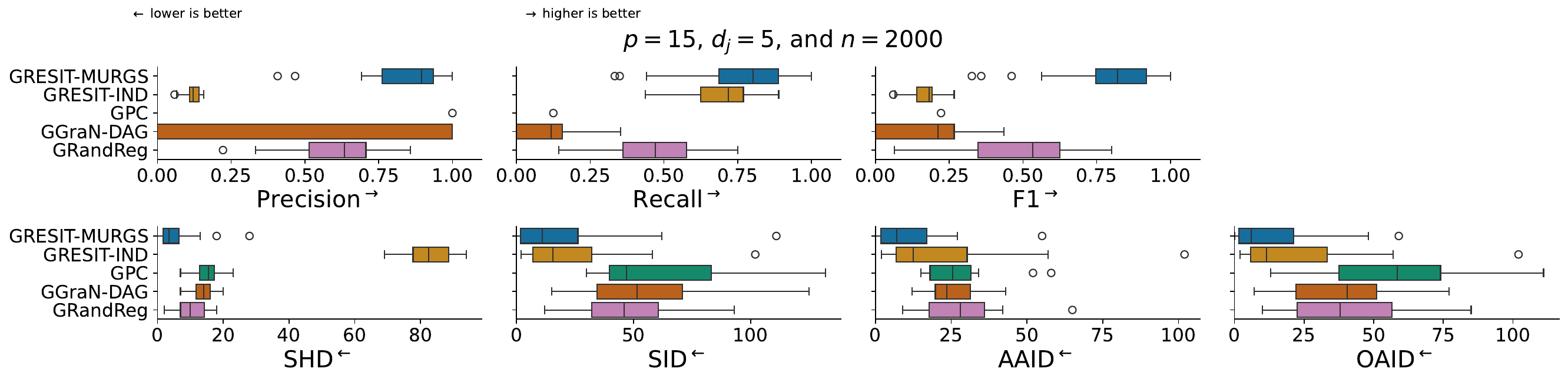}
  \caption{Simulation results based on \(20\) repetitions.}
  \label{fig:sim_results}
\end{figure*}

\paragraph{Results}

Results from our experiments are shown in Figures~\ref{fig:benchmark_scaling} and~\ref{fig:sim_results}. All metrics are averaged over \(20\) independent simulation runs. Synthetic data is generated from GANMs where nonlinear functions are generated from weighted sums of Gaussian processes.

Focusing on AAID, Figure~\ref{fig:benchmark_scaling} illustrates the algorithms' performance across a range of node sizes \(p\), sample sizes \(n\) and group sizes \(d_j\). Across all settings, \textit{GRESIT-MURGS} consistently outperforms \textit{GGraN-DAG}, although it too exhibits challenges when applied to very large graphs with limited sample sizes. The performance difference between \textit{GRESIT-MURGS} and \textit{GGraN-DAG} is particularly pronounced when the group size is varied. Indeed, \textit{GRESIT-MURGS} retains its good performance across different group sizes.

In Figure~\ref{fig:sim_results}, we present a comparison across all considered metrics for two fixed choices of \(p, d_g\), and \(n\). The first row displays classification metrics, where higher values indicate better performance, while the second row shows graph distances, where lower values indicate more accurate graph recovery. In every case, \textit{GRESIT-MURGS} outperforms the other methods. Notably, \textit{GPC} deteriorates as the group size increases, and \textit{GGraN-DAG} suffers similarly, albeit to a lesser extent. Regarding graph distances, the AAID metric is particularly informative, as it reflects the quality of the estimated causal order; here, both GroupRESIT procedures exhibit a clear advantage. In contrast, the pruning phase in \textit{GRESIT-IND} proves ineffective, as evidenced by high SHD.

Overall, the combination of flexible neural networks and nonparametric independence tests proves highly effective in estimating a causal order, provided that the sample size is sufficiently large. Moreover, MURGS demonstrates a robust capability for feature selection even in the presence of large group sizes and numerous candidate parents.

\section{Real manufacturing data application}\label{sec:real_data}

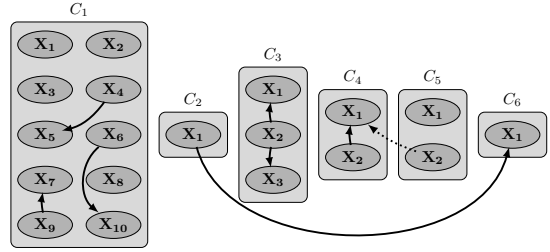
\begin{figure}%[H]
  \centering
  \scalebox{0.6}{
    \tikzset{
  group/.append style={
    align=center,
    text=black,
    scale=1.0
  },
  cell/.append style={
    draw, fill=gray!30!white, rounded corners
  }
}

\begin{tikzpicture}

%C1
\node [cell, minimum width=3cm, minimum height=5cm, label=$C_1$] at (0,0) {};

\draw[fill=gray!60!white] (-0.75, 2) ellipse (0.6cm and 0.3cm) coordinate node [group] (C1_X1) {$\mathbf{X_1}$};
\draw[fill=gray!60!white] (+0.75, 2) ellipse (0.6cm and 0.3cm) coordinate node [group] (C1_X2) {$\mathbf{X_2}$};
\draw[fill=gray!60!white] (-0.75, 1) ellipse (0.6cm and 0.3cm) coordinate node [group] (C1_X3) {$\mathbf{X_3}$};
\draw[fill=gray!60!white] (+0.75, 1) ellipse (0.6cm and 0.3cm) coordinate node [group] (C1_X4) {$\mathbf{X_4}$};
\draw[fill=gray!60!white] (-0.75, 0) ellipse (0.6cm and 0.3cm) coordinate node [group] (C1_X5) {$\mathbf{X_5}$};
\draw[fill=gray!60!white] (+0.75, 0) ellipse (0.6cm and 0.3cm) coordinate node [group] (C1_X6) {$\mathbf{X_6}$};
\draw[fill=gray!60!white] (-0.75, -1) ellipse (0.6cm and 0.3cm) coordinate node [group] (C1_X7) {$\mathbf{X_7}$};
\draw[fill=gray!60!white] (+0.75, -1) ellipse (0.6cm and 0.3cm) coordinate node [group] (C1_X8) {$\mathbf{X_8}$};
\draw[fill=gray!60!white] (-0.75, -2) ellipse (0.6cm and 0.3cm) coordinate node [group] (C1_X9) {$\mathbf{X_9}$};
\draw[fill=gray!60!white] (+0.75, -2) ellipse (0.6cm and 0.3cm) coordinate node [group] (C1_X10) {$\mathbf{X_{10}}$};

\node [cell, minimum width=1.5cm, minimum height=1cm, label=$C_2$] at (2.5,0) {};
\draw[fill=gray!60!white] (2.5, 0) ellipse (0.6cm and 0.3cm) coordinate node [group] (C2_X1) {$\mathbf{X_1}$};

\node [cell, minimum width=1.5cm, minimum height=3cm, label=$C_3$] at (4.25,0) {};
\draw[fill=gray!60!white] (4.25, 1) ellipse (0.6cm and 0.3cm) coordinate node [group] (C3_X1) {$\mathbf{X_1}$};
\draw[fill=gray!60!white] (4.25, 0) ellipse (0.6cm and 0.3cm) coordinate node [group] (C3_X2) {$\mathbf{X_2}$};
\draw[fill=gray!60!white] (4.25, -1) ellipse (0.6cm and 0.3cm) coordinate node [group] (C3_X3) {$\mathbf{X_3}$};

\node [cell, minimum width=1.5cm, minimum height=2cm, label=$C_4$] at (6,0) {};
\draw[fill=gray!60!white] (6, 0.5) ellipse (0.6cm and 0.3cm) coordinate node [group] (C4_X1) {$\mathbf{X_1}$};
\draw[fill=gray!60!white] (6, -0.5) ellipse (0.6cm and 0.3cm) coordinate node [group] (C4_X2) {$\mathbf{X_2}$};

\node [cell, minimum width=1.5cm, minimum height=2cm, label=$C_5$] at (7.75,0) {};
\draw[fill=gray!60!white] (7.75, 0.5) ellipse (0.6cm and 0.3cm) coordinate node [group] (C5_X1) {$\mathbf{X_1}$};
\draw[fill=gray!60!white] (7.75, -0.5) ellipse (0.6cm and 0.3cm) coordinate node [group] (C5_X2) {$\mathbf{X_2}$};

\node [cell, minimum width=1.5cm, minimum height=1cm, label=$C_6$] at (9.5,0) {};
\draw[fill=gray!60!white] (9.5, 0) ellipse (0.6cm and 0.3cm) coordinate node [group] (C6_X1) {$\mathbf{X_1}$};

\draw[-latex, black, very thick] (C1_X4) to[bend left=20] (C1_X5);
\draw[-latex, black, very thick] (C1_X6) to[bend right=50] (C1_X10);
\draw[-latex, black, very thick] (C1_X9) to[bend left=10] (C1_X7);

\draw[-latex, black, very thick] (C2_X1) to[bend right=75] (C6_X1);

\draw[-latex, black, very thick] (C3_X2) to[bend left=10] (C3_X1);
\draw[-latex, black, very thick] (C3_X2) to[bend right=10] (C3_X3);

\draw[-latex, black, very thick] (C4_X2) to[bend left=10] (C4_X1);

\draw[-latex, black, very thick, dotted] (C5_X2) to[bend left=10] (C4_X1);

\end{tikzpicture}
  }
  \caption{Learned causal edges from the real-world dataset using \textit{GRESIT-MURGS}}.
  \label{fig:real_data_comparison}
\end{figure}

We use real-world data from a highly automated production line. Each unit travels along a conveyor
belt and passes through several process cells \(C_1,\ldots,C_6\) (see
Figure~\ref{fig:real_data_comparison}). In each cell, various automated manufacturing processes—such
as staking, welding, etc.—are performed. For example, cell \(C_1\) executes ten distinct tasks,
whereas cell \(C_4\) performs only two. For every process, a set of physical measurements is
extracted and recorded, denoted by \(\mathbf{X}_i\) for cell \(C_i\).
Their dimensionality range from one to \(13\).
In a staking process, for
instance, measurements like the press-in force and the maximum attained force are obtained to ensure
production quality. These individual process measurements form the causal groups of interest. We observe \(19\) groups, which together comprise \(121\) features. The full dataset contains approximately \(35,000\) parts, of which we randomly sample \(5,000\) for computational tractability.
Access to complete causal ground truth in real-world applications is typically prohibitive or unattainable~\citep{Goebler2024}. However, in production lines, partial causal ordering is available through domain knowledge due to the sequential nature of unit flow: processes in cell \(C_i\) can affect only those in cell \(C_j\) with \(i \le j\). Thus, we can partially assess learned graphs by their edge orientation.

Figure~\ref{fig:real_data_comparison} shows edges learned by \textit{GRESIT-MURGS}. Notably, only one edge conflicts with the cell-implied partial order. Conversely, \textit{GPC} yields many undirected edges and two directed edges, one violating the cell arrangement, while \textit{GGraN-DAG} identifies just two edges in cell \(C_3\). Lastly, applying \textit{GroupDirectLiNGAM}\citet{Entner2012} with MURGS pruning (Section\ref{sec:algorithms}) results exclusively in edges violating the partial ordering (see Section~\ref{sec:additional_results}). Thus, \textit{GRESIT-MURGS} produces the most meaningful causal structure in this setting.

\section{Conclusions}\label{sec:conclusions}

In this work, we develop methodology and procedures for nonlinear causal discovery for grouped data. Building upon the RESIT algorithm introduced by \citet{Peters2014}, we propose \textit{MURGS}, a pruning procedure to perform model selection after a causal order has been learned. We derive a closed-form backfitting update within a block coordinate descent framework. Extensive experiments on synthetic data demonstrate that \textit{GRESIT} combined with \textit{MURGS} achieves superior performance, clearly outperforming other state-of-the-art causal discovery algorithms. Moreover, evaluation on real-world manufacturing data with partially known causal structure further substantiates the practical applicability of our method. We have implemented all methods presented in this paper in a Python library, which is publicly available at \url{https://github.com/boschresearch/gresit}.

The current limitations of our work include the assumptions on nonlinearity and differentiability imposed on the function class \(\mathcal{F}\). Next to rather standard assumptions (acyclicity and i.i.d.~data) we assume the absence of unobserved confounding. Addressing these limitations presents promising directions for future work.

Overall, our contributions offer a flexible framework for causal discovery in grouped data under the GANM, laying the groundwork for both theoretical developments and practical applications in complex real-world systems.

% \begin{contributions} % will be removed in pdf for initial submission
%   % (without ‘accepted’ option in \documentclass)
%   % so you can already fill it to test with the
%   % ‘accepted’ class option
%   Briefly list author contributions.
%   This is a nice way of making clear who did what and to give proper credit.
%   This section is optional.

%   H.~Q.~Bovik conceived the idea and wrote the paper.
%   Coauthor One created the code.
%   Coauthor Two created the figures.
% \end{contributions}

\begin{acknowledgements} % will be removed in pdf for initial submission,
  % (without ‘accepted’ option in \documentclass)
  % so you can already fill it to test with the
  % ‘accepted’ class option
  The authors gratefully acknowledge Andrew McCormack, Martin Roth, and Hongjian Shi for their insightful discussions and valuable feedback. We also thank the anonymous referees whose thoughtful suggestions greatly improved the clarity and quality of this work. Mathias Drton acknowledges funding from the German Federal Ministry of Education and Research and the Bavarian State Ministry for Science and the Arts.
\end{acknowledgements}

% References
\bibliography{main}

\begin{thebibliography}{59}
\providecommand{\natexlab}[1]{#1}
\providecommand{\url}[1]{\texttt{#1}}
\expandafter\ifx\csname urlstyle\endcsname\relax
  \providecommand{\doi}[1]{doi: #1}\else
  \providecommand{\doi}{doi: \begingroup \urlstyle{rm}\Url}\fi

\bibitem[Anand et~al.(2023)Anand, Ribeiro, Tian, and Bareinboim]{Anand2023}
Tara~V. Anand, Adele~H. Ribeiro, Jin Tian, and Elias Bareinboim.
\newblock Causal effect identification in cluster dags.
\newblock \emph{Proceedings of the AAAI Conference on Artificial Intelligence}, 37\penalty0 (10):\penalty0 12172--12179, 2023.
\newblock \doi{10.1609/aaai.v37i10.26435}.
\newblock URL \url{https://ojs.aaai.org/index.php/AAAI/article/view/26435}.

\bibitem[Antonoplis(2022)]{Antonoplis2022}
Stephen Antonoplis.
\newblock Studying socioeconomic status: Conceptual problems and an alternative path forward.
\newblock \emph{Perspectives on Psychological Science}, 18\penalty0 (2):\penalty0 275–292, 2022.
\newblock \doi{10.1177/17456916221093615}.
\newblock URL \url{http://dx.doi.org/10.1177/17456916221093615}.

\bibitem[Argyriou et~al.(2006)Argyriou, Evgeniou, and Pontil]{Argyriou2006}
Andreas Argyriou, Theodoros Evgeniou, and Massimiliano Pontil.
\newblock Multi-task feature learning.
\newblock In B.~Sch\"{o}lkopf, J.~Platt, and T.~Hoffman, editors, \emph{Advances in Neural Information Processing Systems}, volume~19. MIT Press, 2006.
\newblock URL \url{https://proceedings.neurips.cc/paper_files/paper/2006/file/0afa92fc0f8a9cf051bf2961b06ac56b-Paper.pdf}.

\bibitem[Bollen(1989)]{Bollen1989}
Kenneth~A. Bollen.
\newblock \emph{Structural equations with latent variables}.
\newblock Wiley Series in Probability and Mathematical Statistics: Applied Probability and Statistics. John Wiley \& Sons, Inc., New York, 1989.
\newblock \doi{10.1002/9781118619179}.
\newblock URL \url{https://doi.org/10.1002/9781118619179}.
\newblock A Wiley-Interscience Publication.

\bibitem[Campbell and Fiske(1959)]{Campbell1959}
Donald~T. Campbell and Donald~W. Fiske.
\newblock Convergent and discriminant validation by the multitrait-multimethod matrix.
\newblock \emph{Psychological Bulletin}, 56\penalty0 (2):\penalty0 81–105, 1959.
\newblock \doi{10.1037/h0046016}.
\newblock URL \url{http://dx.doi.org/10.1037/h0046016}.

\bibitem[Chalupka et~al.(2016)Chalupka, Eberhardt, and Perona]{Chalupka2016}
Krzysztof Chalupka, Frederick Eberhardt, and Pietro Perona.
\newblock Multi-level cause-effect systems.
\newblock In Arthur Gretton and Christian~C. Robert, editors, \emph{Proceedings of the 19th International Conference on Artificial Intelligence and Statistics}, volume~51 of \emph{Proceedings of Machine Learning Research}, pages 361--369, Cadiz, Spain, 2016. PMLR.
\newblock URL \url{https://proceedings.mlr.press/v51/chalupka16.html}.

\bibitem[Chickering(2003)]{Chickering2003}
David~Maxwell Chickering.
\newblock Optimal structure identification with greedy search.
\newblock volume~3, pages 507--554. 2003.
\newblock \doi{10.1162/153244303321897717}.
\newblock URL \url{https://doi.org/10.1162/153244303321897717}.
\newblock Computational learning theory.

\bibitem[Colombo and Maathuis(2014)]{Colombo2014}
Diego Colombo and Marloes~H. Maathuis.
\newblock Order-independent constraint-based causal structure learning.
\newblock \emph{J. Mach. Learn. Res.}, 15:\penalty0 3741--3782, 2014.

\bibitem[Cronbach and Meehl(1955)]{Cronbach1955}
Lee~J. Cronbach and Paul~E. Meehl.
\newblock Construct validity in psychological tests.
\newblock \emph{Psychological Bulletin}, 52\penalty0 (4):\penalty0 281–302, 1955.
\newblock \doi{10.1037/h0040957}.
\newblock URL \url{http://dx.doi.org/10.1037/h0040957}.

\bibitem[Drton and Maathuis(2017)]{Drton2017}
Mathias Drton and Marloes~H. Maathuis.
\newblock Structure learning in graphical modeling.
\newblock \emph{Annual Review of Statistics and Its Application}, 4\penalty0 (1):\penalty0 365--393, 2017.
\newblock \doi{10.1146/annurev-statistics-060116-053803}.
\newblock URL \url{https://doi.org/10.1146/annurev-statistics-060116-053803}.

\bibitem[Entner and Hoyer(2012)]{Entner2012}
Doris Entner and Patrik~O. Hoyer.
\newblock Estimating a causal order among groups of variables in linear models.
\newblock In Alessandro E.~P. Villa, W{\l}odzis{\l}aw Duch, P{\'e}ter {\'E}rdi, Francesco Masulli, and G{\"u}nther Palm, editors, \emph{Artificial Neural Networks and Machine Learning -- ICANN 2012}, pages 84--91, Berlin, Heidelberg, 2012. Springer Berlin Heidelberg.

\bibitem[Erd\"{o}s and Rényi(2011)]{Erdos2011}
P.~Erd\"{o}s and A.~Rényi.
\newblock \emph{On the evolution of random graphs}, pages 38--82.
\newblock Princeton University Press, 2011.
\newblock \doi{10.1515/9781400841356.38}.
\newblock URL \url{http://dx.doi.org/10.1515/9781400841356.38}.

\bibitem[Fornasier and Rauhut(2008)]{Fornasier2008}
Massimo Fornasier and Holger Rauhut.
\newblock Recovery algorithms for vector-valued data with joint sparsity constraints.
\newblock \emph{SIAM Journal on Numerical Analysis}, 46\penalty0 (2):\penalty0 577--613, 2008.
\newblock \doi{10.1137/0606668909}.
\newblock URL \url{https://doi.org/10.1137/0606668909}.

\bibitem[Foygel and Drton(2010)]{Foygel2010}
Rina Foygel and Mathias Drton.
\newblock Exact block-wise optimization in group lasso and sparse group lasso for linear regression, 2010.
\newblock URL \url{https://arxiv.org/abs/1010.3320}.

\bibitem[Fukumizu et~al.(2007)Fukumizu, Gretton, Sun, and Sch\"{o}lkopf]{Fukumizu2007}
Kenji Fukumizu, Arthur Gretton, Xiaohai Sun, and Bernhard Sch\"{o}lkopf.
\newblock Kernel measures of conditional dependence.
\newblock In J.~Platt, D.~Koller, Y.~Singer, and S.~Roweis, editors, \emph{Advances in Neural Information Processing Systems}, volume~20. Curran Associates, Inc., 2007.
\newblock URL \url{https://proceedings.neurips.cc/paper_files/paper/2007/file/3a0772443a0739141292a5429b952fe6-Paper.pdf}.

\bibitem[G\"obler et~al.(2024)G\"obler, Windisch, Drton, Pychynski, Roth, and Sonntag]{Goebler2024}
Konstantin G\"obler, Tobias Windisch, Mathias Drton, Tim Pychynski, Martin Roth, and Steffen Sonntag.
\newblock $\texttt{causalAssembly}$: Generating realistic production data for benchmarking causal discovery.
\newblock In Francesco Locatello and Vanessa Didelez, editors, \emph{Proceedings of the Third Conference on Causal Learning and Reasoning}, volume 236 of \emph{Proceedings of Machine Learning Research}, pages 609--642. PMLR, 2024.
\newblock URL \url{https://proceedings.mlr.press/v236/gobler24a.html}.

\bibitem[Greenfeld and Shalit(2020)]{Greenfeld2020}
Daniel Greenfeld and Uri Shalit.
\newblock Robust learning with the {H}ilbert-schmidt independence criterion.
\newblock In Hal~Daumé III and Aarti Singh, editors, \emph{Proceedings of the 37th International Conference on Machine Learning}, volume 119 of \emph{Proceedings of Machine Learning Research}, pages 3759--3768. PMLR, 2020.
\newblock URL \url{https://proceedings.mlr.press/v119/greenfeld20a.html}.

\bibitem[Gretton et~al.(2005)Gretton, Bousquet, Smola, and Sch\"olkopf]{Gretton2005}
Arthur Gretton, Olivier Bousquet, Alex Smola, and Bernhard Sch\"olkopf.
\newblock Measuring statistical dependence with {H}ilbert-{S}chmidt norms.
\newblock In \emph{Algorithmic learning theory}, volume 3734 of \emph{Lecture Notes in Comput. Sci.}, pages 63--77. Springer, Berlin, 2005.
\newblock \doi{10.1007/11564089\_7}.
\newblock URL \url{https://doi.org/10.1007/11564089_7}.

\bibitem[Hastie et~al.(2015)Hastie, Tibshirani, and Wainwright]{Hastie2015}
Trevor Hastie, Robert Tibshirani, and Martin Wainwright.
\newblock \emph{Statistical learning with sparsity}, volume 143 of \emph{Monographs on Statistics and Applied Probability}.
\newblock CRC Press, Boca Raton, FL, 2015.
\newblock The lasso and generalizations.

\bibitem[Henckel et~al.(2024)Henckel, W{\"u}rtzen, and Weichwald]{Henckel2024}
Leonard Henckel, Theo W{\"u}rtzen, and Sebastian Weichwald.
\newblock Adjustment identification distance: A gadjid for causal structure learning.
\newblock In \emph{The 40th Conference on Uncertainty in Artificial Intelligence}, 2024.
\newblock URL \url{https://openreview.net/forum?id=jO5UNNrjJr}.

\bibitem[Hoyer et~al.(2008)Hoyer, Janzing, Mooij, Peters, and Sch\"{o}lkopf]{Hoyer2009}
Patrik Hoyer, Dominik Janzing, Joris~M Mooij, Jonas Peters, and Bernhard Sch\"{o}lkopf.
\newblock Nonlinear causal discovery with additive noise models.
\newblock In D.~Koller, D.~Schuurmans, Y.~Bengio, and L.~Bottou, editors, \emph{Advances in Neural Information Processing Systems}, volume~21. Curran Associates, Inc., 2008.
\newblock URL \url{https://proceedings.neurips.cc/paper_files/paper/2008/file/f7664060cc52bc6f3d620bcedc94a4b6-Paper.pdf}.

\bibitem[Janzing et~al.(2010)Janzing, Hoyer, and Sch{\"o}lkopf]{Janzing2010}
D.~Janzing, P.~Hoyer, and B.~Sch{\"o}lkopf.
\newblock Telling cause from effect based on high-dimensional observations.
\newblock In \emph{Proceedings of the 27th International Conference on Machine Learning}, pages 479--486, Madison, WI, USA, 2010. Max-Planck-Gesellschaft, International Machine Learning Society.

\bibitem[Kikuchi and Shimizu(2023)]{Kikuchi2023}
Genta Kikuchi and Shohei Shimizu.
\newblock Structure learning for groups of variables in nonlinear time-series data with location-scale noise.
\newblock In Erich Kummerfeld, Sisi Ma, Eric Rawls, and Bryan Andrews, editors, \emph{Proceedings of the 2023 Causal Analysis Workshop Series}, volume 223 of \emph{Proceedings of Machine Learning Research}, pages 20--39. PMLR, 2023.
\newblock URL \url{https://proceedings.mlr.press/v223/kikuchi23a.html}.

\bibitem[Kohn et~al.(2020)Kohn, Jasper, Semedo, Gokcen, Machens, and Yu]{Kohn2020}
Adam Kohn, Anna~I. Jasper, João~D. Semedo, Evren Gokcen, Christian~K. Machens, and Byron~M. Yu.
\newblock Principles of corticocortical communication: Proposed schemes and design considerations.
\newblock \emph{Trends in Neurosciences}, 43\penalty0 (9):\penalty0 725--737, 2020.
\newblock \doi{https://doi.org/10.1016/j.tins.2020.07.001}.
\newblock URL \url{https://www.sciencedirect.com/science/article/pii/S016622362030165X}.

\bibitem[Lachapelle et~al.(2020)Lachapelle, Brouillard, Deleu, and Lacoste-Julien]{Lachapelle2020}
Sébastien Lachapelle, Philippe Brouillard, Tristan Deleu, and Simon Lacoste-Julien.
\newblock Gradient-based neural dag learning.
\newblock In \emph{International Conference on Learning Representations}, 2020.
\newblock URL \url{https://openreview.net/forum?id=rklbKA4YDS}.

\bibitem[Li et~al.(2020)Li, Chang, Han, Zhang, Zaia, and Wan]{Li2020}
Lei Li, Deborah Chang, Lei Han, Xiaojian Zhang, Joseph Zaia, and Xiu-Feng Wan.
\newblock Multi-task learning sparse group lasso: a method for quantifying antigenicity of influenza a(h1n1) virus using mutations and variations in glycosylation of hemagglutinin.
\newblock \emph{BMC Bioinformatics}, 21\penalty0 (1), 2020.
\newblock \doi{10.1186/s12859-020-3527-5}.
\newblock URL \url{http://dx.doi.org/10.1186/s12859-020-3527-5}.

\bibitem[Liu et~al.(2007)Liu, Wasserman, Lafferty, and Ravikumar]{Liu2007}
Han Liu, Larry Wasserman, John Lafferty, and Pradeep Ravikumar.
\newblock Spam: Sparse additive models.
\newblock In J.~Platt, D.~Koller, Y.~Singer, and S.~Roweis, editors, \emph{Advances in Neural Information Processing Systems}, volume~20. Curran Associates, Inc., 2007.
\newblock URL \url{https://proceedings.neurips.cc/paper_files/paper/2007/file/42e7aaa88b48137a16a1acd04ed91125-Paper.pdf}.

\bibitem[Liu et~al.(2008)Liu, Wasserman, and Lafferty]{Liu2008}
Han Liu, Larry Wasserman, and John Lafferty.
\newblock Nonparametric regression and classification with joint sparsity constraints.
\newblock In D.~Koller, D.~Schuurmans, Y.~Bengio, and L.~Bottou, editors, \emph{Advances in Neural Information Processing Systems}, volume~21. Curran Associates, Inc., 2008.
\newblock URL \url{https://proceedings.neurips.cc/paper_files/paper/2008/file/6faa8040da20ef399b63a72d0e4ab575-Paper.pdf}.

\bibitem[Liu et~al.(2009{\natexlab{a}})Liu, Palatucci, and Zhang]{Liu2009a}
Han Liu, Mark Palatucci, and Jian Zhang.
\newblock Blockwise coordinate descent procedures for the multi-task lasso, with applications to neural semantic basis discovery.
\newblock In \emph{Proceedings of the 26th Annual International Conference on Machine Learning}, ICML '09, pages 649--656, 2009{\natexlab{a}}.
\newblock \doi{10.1145/1553374.1553458}.
\newblock URL \url{http://dx.doi.org/10.1145/1553374.1553458}.

\bibitem[Liu et~al.(2009{\natexlab{b}})Liu, Ji, and Ye]{Liu2009}
Jun Liu, Shuiwang Ji, and Jieping Ye.
\newblock Multi-task feature learning via efficient l2, 1-norm minimization.
\newblock In \emph{Proceedings of the Twenty-Fifth Conference on Uncertainty in Artificial Intelligence}, UAI '09, pages 339--348, 2009{\natexlab{b}}.

\bibitem[Mooij et~al.(2009)Mooij, Janzing, Peters, and Sch\"{o}lkopf]{Mooij2009}
Joris Mooij, Dominik Janzing, Jonas Peters, and Bernhard Sch\"{o}lkopf.
\newblock Regression by dependence minimization and its application to causal inference in additive noise models.
\newblock In \emph{Proceedings of the 26th Annual International Conference on Machine Learning}, ICML '09, pages 745--752, 2009.
\newblock \doi{10.1145/1553374.1553470}.
\newblock URL \url{https://doi.org/10.1145/1553374.1553470}.

\bibitem[Negahban et~al.(2012)Negahban, Ravikumar, Wainwright, and Yu]{Negahban2012}
Sahand~N. Negahban, Pradeep Ravikumar, Martin~J. Wainwright, and Bin Yu.
\newblock A unified framework for high-dimensional analysis of {$M$}-estimators with decomposable regularizers.
\newblock \emph{Statist. Sci.}, 27\penalty0 (4):\penalty0 538--557, 2012.
\newblock \doi{10.1214/12-STS400}.
\newblock URL \url{https://doi.org/10.1214/12-STS400}.

\bibitem[Panzeri et~al.(2017)Panzeri, Harvey, Piasini, Latham, and Fellin]{Panzeri2017}
Stefano Panzeri, Christopher~D. Harvey, Eugenio Piasini, Peter~E. Latham, and Tommaso Fellin.
\newblock Cracking the neural code for sensory perception by combining statistics, intervention, and behavior.
\newblock \emph{Neuron}, 93\penalty0 (3):\penalty0 491--507, 2017.
\newblock \doi{https://doi.org/10.1016/j.neuron.2016.12.036}.
\newblock URL \url{https://www.sciencedirect.com/science/article/pii/S0896627316310091}.

\bibitem[Parviainen and Kaski(2017)]{Parviainen2017}
Pekka Parviainen and Samuel Kaski.
\newblock Learning structures of {B}ayesian networks for variable groups.
\newblock \emph{Internat. J. Approx. Reason.}, 88:\penalty0 110--127, 2017.
\newblock \doi{10.1016/j.ijar.2017.05.006}.
\newblock URL \url{https://doi.org/10.1016/j.ijar.2017.05.006}.

\bibitem[Peters and B\"{u}hlmann(2015)]{Peters2015}
Jonas Peters and Peter B\"{u}hlmann.
\newblock Structural intervention distance for evaluating causal graphs.
\newblock \emph{Neural Comput.}, 27\penalty0 (3):\penalty0 771--799, 2015.
\newblock \doi{10.1162/neco\_a\_00708}.
\newblock URL \url{https://doi.org/10.1162/neco_a_00708}.

\bibitem[Peters et~al.(2011)Peters, Mooij, Janzing, and Sch\"{o}lkopf]{Peters2011}
Jonas Peters, Joris~M. Mooij, Dominik Janzing, and Bernhard Sch\"{o}lkopf.
\newblock Identifiability of causal graphs using functional models.
\newblock In \emph{Proceedings of the Twenty-Seventh Conference on Uncertainty in Artificial Intelligence}, UAI'11, pages 589--598, 2011.

\bibitem[Peters et~al.(2014)Peters, Mooij, Janzing, and Sch\"{o}lkopf]{Peters2014}
Jonas Peters, Joris~M. Mooij, Dominik Janzing, and Bernhard Sch\"{o}lkopf.
\newblock Causal discovery with continuous additive noise models.
\newblock \emph{J. Mach. Learn. Res.}, 15:\penalty0 2009--2053, 2014.

\bibitem[Ravikumar et~al.(2009)Ravikumar, Lafferty, Liu, and Wasserman]{Ravikumar2009}
Pradeep Ravikumar, John Lafferty, Han Liu, and Larry Wasserman.
\newblock Sparse additive models.
\newblock \emph{J. R. Stat. Soc. Ser. B Stat. Methodol.}, 71\penalty0 (5):\penalty0 1009--1030, 2009.
\newblock \doi{10.1111/j.1467-9868.2009.00718.x}.
\newblock URL \url{https://doi.org/10.1111/j.1467-9868.2009.00718.x}.

\bibitem[Reisach et~al.(2021)Reisach, Seiler, and Weichwald]{Reisach2021}
{Alexander G.} Reisach, Christof Seiler, and Sebastian Weichwald.
\newblock Beware of the simulated {DAG}! {C}ausal discovery benchmarks may be easy to game.
\newblock In \emph{Advances in Neural Information Processing Systems 34 (NeurIPS)}, pages 1--13. NeurIPS Proceedings, 2021.

\bibitem[Rockafellar and Wets(1998)]{Rockafellar1998}
R.~Tyrrell Rockafellar and Roger J.-B. Wets.
\newblock \emph{Variational analysis}, volume 317 of \emph{Grundlehren der mathematischen Wissenschaften [Fundamental Principles of Mathematical Sciences]}.
\newblock Springer-Verlag, Berlin, 1998.
\newblock \doi{10.1007/978-3-642-02431-3}.
\newblock URL \url{https://doi.org/10.1007/978-3-642-02431-3}.

\bibitem[Rolland et~al.(2022)Rolland, Cevher, Kleindessner, Russell, Janzing, Sch{\"o}lkopf, and Locatello]{Rolland2022}
Paul Rolland, Volkan Cevher, Matth{\"a}us Kleindessner, Chris Russell, Dominik Janzing, Bernhard Sch{\"o}lkopf, and Francesco Locatello.
\newblock Score matching enables causal discovery of nonlinear additive noise models.
\newblock In Kamalika Chaudhuri, Stefanie Jegelka, Le~Song, Csaba Szepesvari, Gang Niu, and Sivan Sabato, editors, \emph{Proceedings of the 39th International Conference on Machine Learning}, volume 162 of \emph{Proceedings of Machine Learning Research}, pages 18741--18753. PMLR, 2022.
\newblock URL \url{https://proceedings.mlr.press/v162/rolland22a.html}.

\bibitem[Rubenstein* et~al.(2017)Rubenstein*, Weichwald*, Bongers, Mooij, Janzing, Grosse-Wentrup, and Sch{\"o}lkopf]{Rubenstein2017}
P.~K. Rubenstein*, S.~Weichwald*, S.~Bongers, J.~M. Mooij, D.~Janzing, M.~Grosse-Wentrup, and B.~Sch{\"o}lkopf.
\newblock Causal consistency of structural equation models.
\newblock In \emph{Proceedings of the 33rd Conference on Uncertainty in Artificial Intelligence (UAI)}, page ID 11, 2017.
\newblock URL \url{http://auai.org/uai2017/proceedings/papers/11.pdf}.
\newblock *equal contribution.

\bibitem[Runge et~al.(2015)Runge, Petoukhov, Donges, Hlinka, Jajcay, Vejmelka, Hartman, Marwan, Paluš, and Kurths]{Runge2015}
Jakob Runge, Vladimir Petoukhov, Jonathan~F. Donges, Jaroslav Hlinka, Nikola Jajcay, Martin Vejmelka, David Hartman, Norbert Marwan, Milan Paluš, and J\"{u}rgen Kurths.
\newblock Identifying causal gateways and mediators in complex spatio-temporal systems.
\newblock \emph{Nature Communications}, 6\penalty0 (1), 2015.
\newblock \doi{10.1038/ncomms9502}.
\newblock URL \url{http://dx.doi.org/10.1038/ncomms9502}.

\bibitem[Semedo et~al.(2020)Semedo, Gokcen, Machens, Kohn, and Yu]{Semedo2020}
João~D Semedo, Evren Gokcen, Christian~K Machens, Adam Kohn, and Byron~M Yu.
\newblock Statistical methods for dissecting interactions between brain areas.
\newblock \emph{Current Opinion in Neurobiology}, 65:\penalty0 59--69, 2020.
\newblock \doi{https://doi.org/10.1016/j.conb.2020.09.009}.
\newblock URL \url{https://www.sciencedirect.com/science/article/pii/S0959438820301367}.
\newblock Whole-brain interactions between neural circuits.

\bibitem[Shimizu et~al.(2006)Shimizu, Hoyer, Hyv\"arinen, and Kerminen]{Shimizu2006}
Shohei Shimizu, Patrik~O. Hoyer, Aapo Hyv\"arinen, and Antti Kerminen.
\newblock A linear non-{G}aussian acyclic model for causal discovery.
\newblock \emph{J. Mach. Learn. Res.}, 7:\penalty0 2003--2030, 2006.

\bibitem[Shimizu et~al.(2011)Shimizu, Inazumi, Sogawa, Hyv{{\"a}}rinen, Kawahara, Washio, Hoyer, and Bollen]{Shimizu2011}
Shohei Shimizu, Takanori Inazumi, Yasuhiro Sogawa, Aapo Hyv{{\"a}}rinen, Yoshinobu Kawahara, Takashi Washio, Patrik~O. Hoyer, and Kenneth Bollen.
\newblock Directlingam: A direct method for learning a linear non-gaussian structural equation model.
\newblock \emph{Journal of Machine Learning Research}, 12\penalty0 (33):\penalty0 1225--1248, 2011.
\newblock URL \url{http://jmlr.org/papers/v12/shimizu11a.html}.

\bibitem[Spirtes et~al.(1993)Spirtes, Glymour, and Scheines]{Spirtes1993}
Peter Spirtes, Clark Glymour, and Richard Scheines.
\newblock \emph{Causation, prediction, and search}, volume~81 of \emph{Lecture Notes in Statistics}.
\newblock Springer-Verlag, New York, 1993.
\newblock \doi{10.1007/978-1-4612-2748-9}.
\newblock URL \url{https://doi.org/10.1007/978-1-4612-2748-9}.

\bibitem[Spirtes et~al.(2000)Spirtes, Glymour, and Scheines]{Spirtes2000}
Peter Spirtes, Clark Glymour, and Richard Scheines.
\newblock \emph{Causation, prediction, and search}.
\newblock Adaptive Computation and Machine Learning. MIT Press, Cambridge, MA, second edition, 2000.
\newblock With additional material by David Heckerman, Christopher Meek, Gregory F. Cooper and Thomas Richardson, A Bradford Book.

\bibitem[Teyssier and Koller(2012)]{Teyssier2012}
Marc Teyssier and Daphne Koller.
\newblock Ordering-based search: A simple and effective algorithm for learning bayesian networks, 2012.
\newblock URL \url{https://arxiv.org/abs/1207.1429}.

\bibitem[Uemura et~al.(2022)Uemura, Takagi, Takayuki, Yoshida, and Shimizu]{Uemura2022}
Kento Uemura, Takuya Takagi, Kambayashi Takayuki, Hiroyuki Yoshida, and Shohei Shimizu.
\newblock A multivariate causal discovery based on post-nonlinear model.
\newblock In Bernhard Schölkopf, Caroline Uhler, and Kun Zhang, editors, \emph{Proceedings of the First Conference on Causal Learning and Reasoning}, volume 177 of \emph{Proceedings of Machine Learning Research}, pages 826--839. PMLR, 2022.
\newblock URL \url{https://proceedings.mlr.press/v177/uemura22a.html}.

\bibitem[Vuković and Thalmann(2022)]{Vukovic2022}
Matej Vuković and Stefan Thalmann.
\newblock Causal discovery in manufacturing: A structured literature review.
\newblock \emph{Journal of Manufacturing and Materials Processing}, 6\penalty0 (1), 2022.
\newblock \doi{10.3390/jmmp6010010}.
\newblock URL \url{https://www.mdpi.com/2504-4494/6/1/10}.

\bibitem[Wahl et~al.(2023)Wahl, Ninad, and Runge]{Wahl2023}
Jonas Wahl, Urmi Ninad, and Jakob Runge.
\newblock Vector causal inference between two groups of variables.
\newblock \emph{Proceedings of the AAAI Conference on Artificial Intelligence}, 37\penalty0 (10):\penalty0 12305--12312, 2023.
\newblock \doi{10.1609/aaai.v37i10.26450}.
\newblock URL \url{https://ojs.aaai.org/index.php/AAAI/article/view/26450}.

\bibitem[Wahl et~al.(2024)Wahl, Ninad, and Runge]{Wahl2024}
Jonas Wahl, Urmi Ninad, and Jakob Runge.
\newblock Foundations of causal discovery on groups of variables.
\newblock \emph{Journal of Causal Inference}, 12\penalty0 (1), 2024.
\newblock \doi{doi:10.1515/jci-2023-0041}.
\newblock URL \url{https://doi.org/10.1515/jci-2023-0041}.

\bibitem[Wang et~al.(2011)Wang, Nie, Huang, Kim, Nho, Risacher, Saykin, Shen, and the Alzheimer's Disease Neuroimaging~Initiative]{Wang2011}
Hua Wang, Feiping Nie, Heng Huang, Sungeun Kim, Kwangsik Nho, Shannon~L. Risacher, Andrew~J. Saykin, Li~Shen, and For the Alzheimer's Disease Neuroimaging~Initiative.
\newblock Identifying quantitative trait loci via group-sparse multitask regression and feature selection: an imaging genetics study of the adni cohort.
\newblock \emph{Bioinformatics}, 28\penalty0 (2):\penalty0 229--237, 2011.
\newblock \doi{10.1093/bioinformatics/btr649}.
\newblock URL \url{https://doi.org/10.1093/bioinformatics/btr649}.

\bibitem[Yin et~al.(2012)Yin, Chen, and Xing]{Yin2012}
Junming Yin, Xi~Chen, and Eric~P Xing.
\newblock Group sparse additive models.
\newblock \emph{Proc. Int. Conf. Mach. Learn.}, 2012:\penalty0 871--878, 2012.

\bibitem[Yuan and Lin(2006)]{Yuan2006}
Ming Yuan and Yi~Lin.
\newblock Model selection and estimation in regression with grouped variables.
\newblock \emph{J. R. Stat. Soc. Ser. B Stat. Methodol.}, 68\penalty0 (1):\penalty0 49--67, 2006.
\newblock \doi{10.1111/j.1467-9868.2005.00532.x}.
\newblock URL \url{https://doi.org/10.1111/j.1467-9868.2005.00532.x}.

\bibitem[Zhang and Hyv\"{a}rinen(2009)]{Zhang2009}
Kun Zhang and Aapo Hyv\"{a}rinen.
\newblock On the identifiability of the post-nonlinear causal model.
\newblock In \emph{Proceedings of the Twenty-Fifth Conference on Uncertainty in Artificial Intelligence}, UAI '09, pages 647--655, 2009.

\bibitem[Zheng et~al.(2018)Zheng, Aragam, Ravikumar, and Xing]{Zheng2018}
Xun Zheng, Bryon Aragam, Pradeep~K Ravikumar, and Eric~P Xing.
\newblock Dags with no tears: Continuous optimization for structure learning.
\newblock In S.~Bengio, H.~Wallach, H.~Larochelle, K.~Grauman, N.~Cesa-Bianchi, and R.~Garnett, editors, \emph{Advances in Neural Information Processing Systems}, volume~31. Curran Associates, Inc., 2018.
\newblock URL \url{https://proceedings.neurips.cc/paper_files/paper/2018/file/e347c51419ffb23ca3fd5050202f9c3d-Paper.pdf}.

\bibitem[Zscheischler et~al.(2011)Zscheischler, Janzing, and Zhang]{Zscheischler2011}
Jakob Zscheischler, Dominik Janzing, and Kun Zhang.
\newblock Testing whether linear equations are causal: a free probability theory approach.
\newblock In \emph{Proceedings of the Twenty-Seventh Conference on Uncertainty in Artificial Intelligence}, UAI'11, pages 839--848, 2011.

\end{thebibliography}

\newpage

\title{Nonlinear Causal Discovery for Grouped Data\\(Supplementary Material)}
\maketitle

\appendix

\section{Proofs of Identifiability}\label{sec:proofs_A}

\subsection{Proof of Theorem~\ref{theorem:bivariate_identifiability}}

\begin{proof}
  If \(\mathcal{G}_0\) is the empty graph we have \(\mathbf{X}_1 \indep \mathbf{X}_2\). If the graph is not empty, and we have \(\mathbf{X}_1 \indep \mathbf{X}_2\) causal minimality is violated. Hence, we assume that the graph is not empty and \(\mathbf{X}_1 \not\indep \mathbf{X}_2\). The joint density has the following form
  \begin{equation*}
    p_{\mathbf{X}_1,\mathbf{X}_2}(\mathbf{x}_1, \mathbf{x}_2) = p_{\mathbf{X}_1}(\mathbf{x}_1)p_{\mathbf{N}_2}({\mathbf{x}_2} - f_2(\mathbf{x}_1)).
  \end{equation*}
  Let us assume \(\mathcal{G}_0\) is not identifiable from \(P_\mathbf{X}\) alone, i.e., there must exist a backward model of the same form
  \begin{equation*}
    p_{\mathbf{X}_1,\mathbf{X}_2}(\mathbf{x}_1, \mathbf{x}_2) = p_{{\mathbf{X}_2}}({\mathbf{x}_2})p_{\mathbf{N}_1}(\mathbf{x}_1-f_1({\mathbf{x}_2})).
  \end{equation*}
  Define
  \begin{equation}\label{eq:forwards_model}
    \pi_1(\mathbf{x}_1,\mathbf{x}_2) \coloneqq \nu({\mathbf{x}_2}-f_2(\mathbf{x}_1)) + \xi(\mathbf{x}_1)
  \end{equation}
  and
  \begin{equation}\label{eq:backwards_model}
    \pi_2(\mathbf{x}_1,\mathbf{x}_2) \coloneqq \tilde\nu(\mathbf{x}_1-f_1({\mathbf{x}_2})) + \eta({\mathbf{x}_2}),
  \end{equation}
  where \(\tilde{\nu} \coloneqq \log p_{\mathbf{N}_1}\) and \(\eta \coloneqq \log p_{\mathbf{X}_2}\). Clearly, we have that \(\pi_1(\mathbf{x}_1,\mathbf{x}_2) = \pi_2(\mathbf{x}_1,\mathbf{x}_2) = \log p_{\mathbf{X}_1,\mathbf{X}_2}(\mathbf{x}_1,\mathbf{x}_2)\).

  Considering first the backwards model (Eq.~\eqref{eq:backwards_model}) we derive the gradient \(\nabla_{\mathbf{x}_1} \pi_2(\mathbf{x}_1,\mathbf{x}_2)\) with respect to \(\mathbf{x}\), i.e.,
  \begin{equation}
    \nabla_{\mathbf{x}_1} \pi_2(\mathbf{x}_1,\mathbf{x}_2) = \nabla \tilde\nu(\mathbf{u}),
  \end{equation}
  where \(\tilde{\mathbf{u}} \coloneqq \tilde{\mathbf{u}}(\mathbf{x}_1,\mathbf{x}_2) \coloneqq \mathbf{x}_1- f_1({\mathbf{x}_2})\).
  Then, the second derivatives take the following form
  \begin{equation}
    D_{\mathbf{x}_1\mathbf{x}_2}\pi_2(\mathbf{x}_1,\mathbf{x}_2) = -\mathbf{J}_{f_1}({\mathbf{X}_2})^\top \mathbf{H}_{\tilde\nu}(\tilde{\mathbf{u}}),
  \end{equation}
  and
  \begin{equation}
    D_{\mathbf{x}_1\mathbf{x}_1} \pi_2(\mathbf{x}_1,\mathbf{x}_2) = \mathbf{H}_{\tilde\nu}(\tilde{\mathbf{u}}),
  \end{equation}
  with Jacobian \(\mathbf{J}_{f_1} \in \mathbb{R}^{d_{x_1} \times d_{x_2}}\) and Hessian \(\mathbf{H}_{\tilde\nu} \in \mathbb{R}^{d_{x_1} \times d_{x_1}}\). Since we have assumed that \(f_1\) and \(f_2\) are three times continuously differentiable the Hessian is symmetric (Schwarz's theorem) and invertible such that
  \begin{equation*}
    \begin{split}
      {Q}_2(\mathbf{x}_1,\mathbf{x}_2) &\coloneqq \left(D_{\mathbf{xx}} \pi_2(\mathbf{x}_1,\mathbf{x}_2)\right)^{-1} \left(D_{\mathbf{x}_1\mathbf{x}_2}\pi_2(\mathbf{x}_1,\mathbf{x}_2)\right)^\top  \\
      &= \mathbf{H}_{\tilde\nu}(\tilde{\mathbf{u}})^{-1}(-\mathbf{J}_{f_1}({\mathbf{x}_2})^\top \mathbf{H}_{\tilde\nu}(\tilde{\mathbf{u}}))^{\top} \\
      &= -\mathbf{J}_{f_1}({\mathbf{x}_2}),
    \end{split}
  \end{equation*}
  which does not depend on \(\mathbf{x}_1\).

  Now, we repeat the above steps for Eq.~\eqref{eq:forwards_model}, i.e.,
  \begin{equation}
    \nabla_{\mathbf{x}_1} \pi_1(\mathbf{x}_1,\mathbf{x}_2) = -\mathbf{J}_{f_2}(\mathbf{x}_1)^\top \nabla \nu\bigl(\mathbf{u}\bigr) + \nabla \xi(\mathbf{x}_1),
  \end{equation}
  and
  \begin{equation}
    D_{\mathbf{x}_1\mathbf{x}_2}\pi_1(\mathbf{x}_1,\mathbf{x}_2) = -\mathbf{J}_{f_2}(\mathbf{x}_1)^\top \mathbf{H}_{\nu}\bigl(\mathbf{u}\bigr),
  \end{equation}
  where \(\mathbf{J}_{f_2}(\mathbf{x}_1) \in \mathbb{R}^{d_{x_2} \times d_{x_1}}\), \(\nabla \xi(\mathbf{x}_1) \in \mathbb{R}^{d_{x_1}}\), and \(\mathbf{H}_{\nu}\bigl(\mathbf{u}\bigr) \in \mathbb{R}^{d_{x_2} \times d_{x_2}}\). Finally,
  \begin{align}
    D_{\mathbf{x}_1\mathbf{x}_1}\pi_1(\mathbf{x}_1,\mathbf{x}_2) &= \mathbf{H}_{\xi}(\mathbf{x}_1)
    - \mathbf{H}_{f_2}(\mathbf{x}_1)[\nabla \nu\bigl(\mathbf{u}\bigr)] \\
    &+ \mathbf{J}_{f_2}(\mathbf{x}_1)^\top \mathbf{H}_{\nu}\bigl(\mathbf{u}\bigr) \mathbf{J}_{f_2}(\mathbf{x}_1),
  \end{align}
  where \(\mathbf{H}_{\xi}(\mathbf{x}_1) \in \mathbb{R}^{d_{x_1} \times d_{x_1}}\) and the Hessian \(\mathbf{H}_{f_2} \in \mathbb{R}^{d_{x_2} \times d_{x_1} \times d_{x_1}}\) is a third-order tensor such that the contraction with the vector \(\nabla \nu\bigl(\mathbf{u}\bigr)\) may be written as \(\mathbf{H}_{f_2}[\nabla \nu\bigl(\mathbf{u}\bigr)]\), i.e.
  \begin{equation*}
    \left( \mathbf{H}_{f_2}[\nabla \nu\bigl(\mathbf{u}\bigr)] \right)_{ik} \\
    =\sum_{j=1}^{d_{x_2}} \frac{\partial^2 f_j(\mathbf{x}_1)}{\partial x_{1i} \partial x_{1k}}\,\left[\nabla \nu\bigl(\mathbf{u}\bigr)\right]_j.
  \end{equation*}
  We find
  \begin{align*}
    Q_1(\mathbf{x}_1, \mathbf{x}_2) &\coloneqq (D_{\mathbf{x}_1\mathbf{x}_1}\pi_1(\mathbf{x}_1,\mathbf{x}_2))^{-1}(D_{\mathbf{x}_1\mathbf{x}_2}\pi_1(\mathbf{x}_1,\mathbf{x}_2)) \\
    &= \bigg[\mathbf{H}_{\xi}(\mathbf{x}_1)
      - \mathbf{H}_{f_2}(\mathbf{x}_1)[\nabla \nu(\mathbf{u})]\\
    &+ \mathbf{J}_{f_2}(\mathbf{x}_1)^\top \mathbf{H}_{\nu}\bigl(\mathbf{u}\bigr) \mathbf{J}_{f_2}(\mathbf{x}_1)\bigg]^{-1} \\
    &\ [-\mathbf{J}_{f_2}(\mathbf{x}_1)^\top \mathbf{H}_{\nu}(\mathbf{u})].
  \end{align*}

  Recall that \(Q_2(\mathbf{x}_1, \mathbf{x}_2) = Q_2(\mathbf{x}_2)\) is essentially a function only of \({\mathbf{x}_2}\) such that \(D_{\mathbf{x}_1} Q_2(\mathbf{x}_1, \mathbf{x}_2) = \mathbf{0} \in \mathbb{R}^{d_{x_1}\times d_{x_1} \times d_{x_2}}\). Since \(Q_1(\mathbf{x}_1, \mathbf{x}_2) = Q_2(\mathbf{x}_1, \mathbf{x}_2)\), taking the derivative with respect to \(\mathbf{x}_1\) yields
  \begin{align*}
    D_{\mathbf{x}_1}  Q_1(\mathbf{x}_1, \mathbf{x}_2) &= D_{\mathbf{x}_1}  \big[(D_{\mathbf{x}_1\mathbf{x}_1}\pi_1(\mathbf{x}_1,\mathbf{x}_2))^{-1} \\
    & \quad (D_{\mathbf{x}_1\mathbf{x}_2}\pi_1(\mathbf{x}_1,\mathbf{x}_2))\big] \\
    &= \mathbf{0} \in \mathbb{R}^{d_{x_1}\times d_{x_1} \times d_{x_2}}.
  \end{align*}
  Note that each slice \(\partial Q_1(\mathbf{x}_1, \mathbf{x}_2)/\partial x_{1k}\) is matrix-valued, and \(d_{x_1}\)-many such slices exist. Thus, we apply the product rule component-wise for each slice of the tensor and subsequently stack them together, i.e.
  \begin{align*}
    D_{\mathbf{x}_1} Q_1(\mathbf{x}_1, \mathbf{x}_2) &= D_{\mathbf{x}_1}(D_{\mathbf{x}_1\mathbf{x}_1}\pi_1(\mathbf{x}_1,\mathbf{x}_2)^{-1})\\
    &\quad D_{\mathbf{x}_1\mathbf{x}_2}\pi_1(\mathbf{x}_1,\mathbf{x}_2) \\
    &+ (D_{\mathbf{x}_1\mathbf{x}_1}\pi_1(\mathbf{x}_1,\mathbf{x}_2))^{-1} \\
    &\quad D_{\mathbf{x}_1}(D_{\mathbf{x}_1\mathbf{x}_2}\pi_1(\mathbf{x}_1,\mathbf{x}_2)).
  \end{align*}
  Using the identity
  \begin{equation*}
    D_{\mathbf{x}_1}(\mathbf{A})^{-1}
    = - (\mathbf{A})^{-1}
    D_{\mathbf{x}_1}\mathbf{A} (\mathbf{A})^{-1},
  \end{equation*}
  where \(\mathbf{A} \coloneqq \mathbf{A}(\mathbf{x}_1,\mathbf{x}_2) \in \mathbb{R}^{d_{x_1}\times d_{x_1}}\). Therefore, we obtain
  \begin{multline*}
    -(D_{\mathbf{x}_1\mathbf{x}_1}\pi_1)^{-1}
    D_{\mathbf{x}_1}(D_{\mathbf{x}_1\mathbf{x}_1}\pi_1) (D_{\mathbf{x}_1\mathbf{x}_1}\pi_1)^{-1} D_{\mathbf{x}_1\mathbf{x}_2}\pi_1 \\
    = -(D_{\mathbf{x}_1\mathbf{x}_1}\pi_1)^{-1}
    D_{\mathbf{x}_1} (D_{\mathbf{x}_1\mathbf{x}_2}\pi_1),
  \end{multline*}
  where we drop the arguments of \(\pi_1(\mathbf{x}_1,\mathbf{x}_2)\), i.e., \(\pi_1 \coloneqq \pi_1(\mathbf{x}_1,\mathbf{x}_2)\) to improve readability.
  Making sure that all of the matrix-tensor products act on the appropriate indices of the tensors, the expression simplifies to
  \begin{equation*}
    D_{\mathbf{x}_1}(D_{\mathbf{x}_1\mathbf{x}_1}\pi_1) (D_{\mathbf{x}_1\mathbf{x}_1}\pi_1)^{-1} D_{\mathbf{x}_1\mathbf{x}_2}\pi_1 = D_{\mathbf{x}_1} (D_{\mathbf{x}_1\mathbf{x}_2}\pi_1).
  \end{equation*}
  Note that only \(D_{\mathbf{x}_1}(D_{\mathbf{x}_1\mathbf{x}_1}\pi_1)\) contains third order derivatives of the log marginal \(\xi\). Let us state this more explicitly.
  \begin{align*}
    D_{\mathbf{x}_1}(D_{\mathbf{x}_1\mathbf{x}_1}\pi_1) &= D_{\mathbf{x}_1}\Big[\mathbf{H}_\xi(\mathbf{x}_1) - \mathbf{H}_{f_2}(\mathbf{x}_1)[\nabla \nu(\mathbf{u})] \\
    &+ \mathbf{J}_{f_2}(\mathbf{x}_1)^\top \mathbf{H}_{\nu}(\mathbf{u}) \mathbf{J}_{f_2}(\mathbf{x}_1) \Big] \\
    &= D_{\mathbf{x}_1}\mathbf{H}_\xi(\mathbf{x}_1) - D_{\mathbf{x}_1}(\mathbf{H}_{f_2}(\mathbf{x}_1)[\nabla \nu(\mathbf{u})]) \\
    &+ D_{\mathbf{x}_1}(\mathbf{J}_{f_2}(\mathbf{x}_1)^\top \mathbf{H}_{\nu}(\mathbf{u}) \mathbf{J}_{f_2}(\mathbf{x}_1)).
  \end{align*}
  And finally we have
  \begin{multline}\label{app_eq:tensor_differential_eq}
    D_{\mathbf{x}_1}\mathbf{H}_\xi(\mathbf{x}_1) (D_{\mathbf{x}_1\mathbf{x}_1}\pi_1)^{-1} D_{\mathbf{x}_1\mathbf{x}_2}\pi_1 \\
    \begin{aligned}
      = &D_{\mathbf{x}_1} D_{\mathbf{x}_1\mathbf{x}_2}\pi_1 \Big[
        D_{\mathbf{x}_1}\big( \mathbf{H}_{f_2}(\mathbf{x}_1)[\nabla \nu(\mathbf{u})] \big) \\
        - &D_{\mathbf{x}_1}\big( \mathbf{J}_{f_2}(\mathbf{x}_1)^\top \mathbf{H}_{\nu}(\mathbf{u}) \mathbf{J}_{f_2}(\mathbf{x}_1) \big)
      \Big] \\
      (&D_{\mathbf{x}_1\mathbf{x}_1}\pi_1)^{-1} D_{\mathbf{x}_1\mathbf{x}_2}\pi_1
    \end{aligned}
  \end{multline}
  where the remaining second order derivatives of the log marginal \(\xi\) are contained in the expression for \(D_{\mathbf{x}_1\mathbf{x}_1}\pi_1\).
  This contradicts the assumption that \(P_\mathbf{X}\) is generated from a \emph{identifiable bivariate} GANM.
\end{proof}

\subsection{Proof of Corollary~\ref{corollary:multivariate_identifiability}}

\begin{proof}
  Suppose there are two identifiable GANMs that both induce the distribution \(P_\mathbf{X}\) with DAGs \(\mathcal{G}_0\) and \(\mathcal{G}_0^\prime\), respectively. For any two groups \(\mathbf{X}_Q, \mathbf{X}_R\) that satisfy Proposition 29 in~\citet{Peters2014} we consider the set of parents without \(R\) in \(\mathcal{G}_0\), i.e, \({pa}_Q \coloneqq pa_{\mathcal{G}_0}({Q}) \setminus R\) and the set of parents without \(Q\) in \(\mathcal{G}_0^\prime\), i.e, \({pa}_R \coloneqq pa_{\mathcal{G}_0^\prime}({R}) \setminus Q\). Denote their union by \(S \coloneqq {pa}_Q \cup {pa}_R\). For any \(s = ({q},{r})\) we write \(\mathbf{X}^*_{Q} = \mathbf{X}_{Q} \mid_{S=s}\) and \(\mathbf{X}^*_{R} = \mathbf{X}_{R} \mid_{R=r}\). From Lemma~\ref{lemma:ancestor_independence}, we have that \(\mathbf{N}_Q \indep \mathbf{X}_R, \mathbf{X}_S\) and \(\mathbf{N}_R \indep \mathbf{X}_Q, \mathbf{X}_S\). Thus, by Lemma 2 in \citet{Peters2011}, we have the following bivariate GANM in \(\mathcal{G}_0\)
  \begin{equation*}
    \mathbf{X}^*_{Q} = f_{Q}(\mathbf{X}_q, \mathbf{X}^*_{R}) + \mathbf{N}_Q, \quad \mathbf{N}_Q \indep \mathbf{X}^*_{R},
  \end{equation*}
  and in \(\mathcal{G}_0^\prime\)
  \begin{equation*}
    \mathbf{X}^*_{R} = f_{R}(\mathbf{X}_r, \mathbf{X}^*_{Q}) + \mathbf{N}_R, \quad \mathbf{N}_R \indep \mathbf{X}^*_{Q},
  \end{equation*}
  However this is a contradiction since in Corollary~\ref{corollary:multivariate_identifiability} we can choose any \(s=(q,r)\) that ensures that the bivariate GANMs are identifiable.
\end{proof}

\subsection{Proof of Lemma~\ref{lemma:ancestor_independence}}

\begin{proof}
  Write \(\mathbf{S} = (S_1, \dots, S_k)\) such that \(\mathbf{S} = (f_{S_1}(\mathbf{X}_{pa(S_1)}, \mathbf{N}_{S_1}), \ldots, f_{S_k}(\mathbf{X}_{pa(S_k)}, \mathbf{N}_{S_k}) )\). It takes finitely many steps to recursively substitute the parents of \(S_i, i \in [k]\) by the corresponding structural equation such that \(\mathbf{S} = f(\mathbf{N}_{A_1}, \ldots, \mathbf{N}_{A_l})\) with \(\{A_1, \ldots, A_l\}\) the set of all ancestors of nodes in \(\mathbf{S}\) that do not contain the node \(g\). The statement follows from the joint independence of the noise variables across groups.
\end{proof}

\section{Proof of Backfitting Update}\label{sec:backfitting_update}

\begin{lemma}\label{lemma:stationary_cond}
  The stationary condition of Problem~\eqref{eq:backfitting} with respect to \(\mathbf{f}_g\) is given by
  \begin{equation}
    f_{g,h}^{(k)} + \sum_{h' \in [d_g]: h' \neq h} P_hf_{g,h'}^{(k)} - P_hR_g^{(k)} + \lambda\sqrt{d_g}u^{(k)}v_{h}^{(k)} = 0,
  \end{equation}
  for all \(h \in [d_g]\) where \(u^{(k)}\) are scalars and \(\mathbf{v}_g^{(k)} = (v_{h}^{(k)})_{h\in[d_g]}\) is a vector of measurable functions of \(X_h^{(g)}\), with
  \begin{equation}
    (u^{(1)}, \dots, u^{(d_j)})^T \in \partial \norm{\cdot}_\infty\rvert_{(\norm{\mathbf{f}_g^{(1)}}, \dots, \norm{\mathbf{f}_g^{(d_j)}})^T},
  \end{equation}
  and
  \begin{equation}
    \mathbf{v}_{g}^{(k)} \in \partial \norm{\mathbf{f}_g^{(k)}},
  \end{equation}
  for \(k = 1, \dots {d_j}\). The subdifferential of the sup-norm evaluated at \((\norm{\mathbf{f}_g^{(1)}}, \dots, \norm{\mathbf{f}_g^{(d_j)}})^T\) lies in the \(d_j\)-dimensional Euclidean space.
\end{lemma}

\begin{proof}
  Since both the loss function and the regularization term are convex, the solution to the objective function in \eqref{eq:backfitting} can be characterized by the Karush-Kuhn-Tucker conditions. We investigate the subdifferential for the loss and the regularization term separately. For readability, we omit the argument of the component function when it is clear from the context. Starting with the loss function we define
  \begin{equation*}
    L(\mathbf{f}_g^{(k)}) \coloneq \frac{1}{2} \mathbb{E}\left[ \sum_{k=1}^{d_j} \bigg(R_g^{(k)} - \sum_{h\in [d_g]} f_{g,h}^{(k)}\bigg)^2 \right].
  \end{equation*}
  Consider a perturbation of \(L(\mathbf{f}_g^{(k)})\) along the direction
  \begin{equation*}
    \mathbf{\psi}_g^{(k)} = \left(\psi_{g,h}^{(k)} \in \mathcal{H}_{g,h}^{(k)}\right)_{h \in [d_g]}
  \end{equation*}
  such that
  \begin{multline*}
    \lim_{\tau \to 0} \frac{L(\mathbf{f}_g^{(k)} + \tau\mathbf{\psi}_g^{(k)}) - L(\mathbf{f}_g^{(k)})}{\tau} \\
    \begin{aligned}
      &= \sum_{h\in [d_g]}\mathbb{E}\Big[ (\sum_{h' \in [d_g]} f_{g,h'}^{(k)} -R_g^{(k)})\psi_{g,h}^{(k)} \Big] \\
      &= \sum_{h\in [d_g]}\mathbb{E}\Bigg[ \mathbb{E}\Big[\sum_{h' \in [d_g]} f_{g,h'}^{(k)} -R_g^{(k)} \mid X_h^{(g)}\Big]\psi_{g,h}^{(k)} \Bigg] \\
      &= \sum_{h\in [d_g]} \bigg\langle \mathbb{E}\Big[\sum_{h' \in [d_g]} f_{g,h'}^{(k)} -R_g^{(k)} \mid X_h^{(g)}\Big], \psi_{g,h}^{(k)} \bigg\rangle.
    \end{aligned}
  \end{multline*}
  The second equation follows from the law of iterated expectation and the third from expressing the resulting expectation as an inner product in the Hilbert space \(\mathcal{H}_{g,h}^{(k)}\). The gradient of \(L(\mathbf{f}_g^{(k)})\) is
  \begin{equation*}
    \nabla L(\mathbf{f}_g^{(k)}) = \left[\mathbb{E}\Big[\sum_{h' \in [d_g]} f_{g,h'}^{(k)} -R_g^{(k)} \mid X_h^{(g)}\Big]\right]_{h\in [d_g]}.
  \end{equation*}
  The subdifferential of \(\Phi^{d_j}_{\text{group}}(f)\) is given by
  \begin{equation*}
    \partial \Phi^{d_j}_{\text{group}}(f) = \lambda \sqrt{d_g} u_{hk} v_{hk}, \quad \forall h \in [d_g],
  \end{equation*}
  where \((u_{h1}, \dots, u_{h{d_j}})^T \in \partial \norm{\cdot}_\infty\rvert_{(\norm{\mathbf{f}_g^{(1)}}, \dots, \norm{\mathbf{f}_g^{(d_j)}})^T}\) and \(v_{hk} \in \partial \norm{\mathbf{f}_g^{(k)}}\).

  Isolating \(f_{g,h}^{(k)}(X_h^{(g)}) = \mathbb{E}[f_{g,h}^{(k)}(X_h^{(g)}) \mid X_h^{(g)}]\) and using the conditional expectation operator \(\mathbb{E}[\ \cdot \mid X_h^{(g)}]\) for the remaining terms yields the expression for the stationary condition.
\end{proof}

The following two Lemmas characterize the subdifferential of sup-norms (Lemma~\ref{lemma:sup_norm}, proof is provided in \citep[Chapter 8]{Rockafellar1998}) and that of the Euclidean norm (Lemma~\ref{lemma:euclidean}).

\begin{lemma}\label{lemma:sup_norm}
  The subdifferential of \(\norm{\cdot}_\infty\) in \(\R^{d_j}\) is
  \begin{equation}
    \partial \norm{\cdot}_\infty \rvert_x =
    \begin{cases}
      \{\eta : \norm{\eta}_1 \leq 1\}                                   & \text{if } \bm{x} = \bm{0} \\
      \text{conv}\{\text{sign}(x_k) e_k : \abs{x_k} = \norm{x}_\infty\} & \text{o.w.},
    \end{cases}
  \end{equation}
  where \(\text{conv}(A)\) denotes the convex hull of set \(A\) and \(e_k\) is the \(k^{th}\) canonical unit vector in \(\R^{d_j}\).
\end{lemma}

\begin{lemma}\label{lemma:euclidean}
  The subdifferential of \(\norm{\bf{f}_g}\) is
  \begin{equation}
    \partial \norm{f_g} =
    \begin{cases}
      \{f_j / \norm{\bf{f}_g}\}_{j\in g}    & \text{if } \norm{\bf{f}_g} \neq 0 \\
      \{\bf{v}_g : \norm{\bf{v}_g} \leq 1\} & \text{if } \norm{\bf{f}_g} = 0
    \end{cases}
  \end{equation}
\end{lemma}

The proof proceeds by considering three cases for the sup-norm subdifferential evaluated at \((\norm{\mathbf{f}_g^{(1)}}, \dots, \norm{\mathbf{f}_g^{(d_j)}})^T\): (1)
\(\norm{\mathbf{f}_g^{(k)}} = 0\) for all \(k = 1, \dots, d_j\); (2) there exists a unique \(k\), such that \(\norm{\mathbf{f}_g^{(k)}} = \max_{k' = 1, \dots, d_j} \norm{\mathbf{f}_g^{(k')}}\); (3) There exist at least two \(k \neq k'\), such that \(\norm{\mathbf{f}_g^{(k)}} = \norm{\mathbf{f}_g^{(k')}} = \max_{m = 1, \dots, d_j} \norm{\mathbf{f}_g^{(m)}}\)

We begin with the proof of Proposition~\ref{prop:all_zeros}, i.e., the case where \(\sum_{k=1}^{d_j} \norm{\mathbf{Q}R_g^{(k)}} \leq \lambda \sqrt{d_g}\) and show that \(\norm{\mathbf{f}_g^{(k)}} = 0\) must be a solution.

\begin{proof}
  From Lemma~\ref{lemma:stationary_cond} we know that if \(\norm{\mathbf{f}_g^{(k)}} = 0\) then \(\norm{\mathbf{u}}_1 \leq 1\) and \(\norm{\mathbf{v}_{g}^{(k)}} \leq 1\). It follows that
  \begin{align*}
    P_hR_g^{(k)} &= \lambda\sqrt{d_g}u^{(k)}v_{h}^{(k)} \\
    \sum_{k=1}^{{d_j}} \sqrt{\sum_{h\in [d_g]} \mathbb{E}[(P_h R_g^{(k)})^2]} &\leq \lambda \sqrt{d_g}.
  \end{align*}
  On the other hand, we also know from Lemma~\ref{lemma:stationary_cond} that \(\norm{\mathbf{f}_g^{(k)}} = 0\) if and only if \(\exists u^{(1)}, \dots, u^{(d_j)}\) such that \(\sum_{k=1}^{d_j} \abs{u}^{(k)} \leq 1\) and \(\exists v_{1}^{(k)}, \ldots, v_{d_g}^{(k)}\) such that \(\sqrt{\sum_{h=1}^{d_g} (v_h^{2})^{(k)}} \leq 1\) and
  \begin{equation*}
    \lambda\sqrt{d_g}u^{(k)}v_{h}^{(k)} = P_hR_g^{(k)}.
  \end{equation*}
  Then, if \(\sum_{k=1}^{d_j} \norm{\mathbf{Q}R_g^{(k)}} \leq \lambda \sqrt{d_g}\), choosing \(u^{(k)}\) and \(v_{h}^{(k)}\) as above guarantees that \(\sum_{k=1}^{d_j} \abs{u}^{(k)} \leq 1\) and \(\sum_{k=1}^{d_j} \abs{u}^{(k)} \leq 1\), therefore \(\norm{\mathbf{f}_g^{(k)}} = 0\).
\end{proof}

If we continue to allow the within-group covariances to be nonzero, we obtain the following known result.

\begin{lemma}
  Consider the case where there exists a unique \(k\) at which the sup-norm is attained. Then the group SpAM by \citep{Yin2012} is recovered.
\end{lemma}

\begin{proof}
  Note that we must have that \(\sum_{k=1}^{d_j} \norm{\mathbf{Q}R_g^{(k)}} > \lambda \sqrt{d_g}\) otherwise all \(f_{g,h}^{(k)} = 0, \forall h \in [d_g] ,k \in [d_j]\).

  Denote \(k_1 \in [d_j]\) the unique \(k\) that attains the sup-norm, then \(\norm{\mathbf{f}_g^{(k_1)}} > \norm{\mathbf{f}_g^{(k)}}\) for all \(k \neq k_1\). Consequently, the subdifferential of the sup-norm becomes \(\partial \norm{\cdot}_\infty\rvert_{(\norm{\mathbf{f}_g^{(1)}}, \dots, \norm{\mathbf{f}_g^{(d_j)}})^T} = e_{k_1}\), the \(k_1\)-th canonical vector in \(\R^{d_j}\). Hence, from Lemma~\ref{lemma:stationary_cond} we have that
  \begin{multline}\label{eq:case_2_kkt}
    P_hR_g^{(k)} - \Big(f_{g,h}^{(k)} + \sum_{h' \in [d_g]: h' \neq h} P_hf_{g,h'}^{(k)}\Big) \\
    = \lambda\sqrt{d_g}\frac{f_{g,h}^{(k)}}{\norm{\mathbf{f}_{g}^{(k)}}} \mathbbm{1}_{\{k = k_1\}}.
  \end{multline}
  If \(k \neq k_1\) then Equation~\eqref{eq:case_2_kkt} implies
  \begin{equation*}
    \mathbf{f}_g^{(k)} = \mathbf{\mathcal{I}}^{-1} \mathbf{Q}R_g^{(k)},
  \end{equation*}
  where
  \begin{equation*}
    \mathbf{\mathcal{I}} =
    \begin{bmatrix}
      1 & P_1 & \cdots & P_1 \\
      P_2 & 1 & \cdots & P_2 \\
      \vdots & \vdots & \ddots  &  \vdots \\
      P_{d_g} & P_{d_g} & \cdots  & 1
    \end{bmatrix}
  \end{equation*}

  On the other hand, if \(k = k_1\), we have
  \begin{align*}
    \mathbf{Q}R_g^{(k_1)} - \mathbf{\mathcal{I}}\mathbf{f}_g^{(k_1)} &= \lambda\sqrt{d_g}\frac{\mathbf{f}_{g}^{(k_1)}}{\norm{\mathbf{f}_{g}^{(k_1)}}} \\
    \Bigg[\mathbf{\mathcal{I}} + \frac{\lambda\sqrt{d_g}}{\norm{\mathbf{f}_{g}^{(k_1)}}} I_{d_g}\Bigg] \mathbf{f}_g^{(k_1)} &= \mathbf{Q}R_g^{(k_1)},
  \end{align*}
  from which we obtain the group SpAM result discussed by \citet{Yin2012}, i.e.,
  \begin{equation*}
    \mathbf{f}_g^{(k_1)} = \Bigg[\mathbf{\mathcal{I}} + \frac{\lambda\sqrt{d_g}}{\norm{\mathbf{f}_{g}^{(k_1)}}}I_{d_g}\Bigg]^{-1}\mathbf{Q}R_g^{(k_1)}.
  \end{equation*}

\end{proof}

For the remainder of the backfitting update derivation, assume that
\begin{equation*}
  P_hf_{g,h'}^{(k)} = \mathbb{E}[f_{g,h'}^{(k)} \mid X_{h}^{(g)}] =  0, \quad \forall h' \neq h.
\end{equation*}
This implies that the covariance of the within-group component functions is zero, i.e.,
\begin{align*}
  Cov(f_{g,h}^{(k)}, f_{g,h'}^{(k)}) &= \mathbb{E}[f_{g,h}^{(k)} f_{g,h'}^{(k)}] \\
  &= \mathbb{E}[f_{g,h}^{(k)} \mathbb{E}[f_{g,h'}^{(k)} \mid X_{h}^{(g)}]] \\
  &= 0.
\end{align*}

Under this assumption, the stationary condition in Lemma~\ref{lemma:stationary_cond} simplifies to
\begin{equation*}
  f_{g,h}^{(k)} - P_hR_g^{(k)} + \lambda\sqrt{d_g}u^{(k)}v_{h}^{(k)} = 0.
\end{equation*}

Suppose \(k_1 \in [d_j]\) is the unique \(k\) that attains the sup-norm (case 2). Then, we have the simplified expression
\begin{equation}\label{eq:case_2_not_k_1}
  {f}_{g,h}^{(k)} = {P}_{h}^{(k)}R_g^{(k)}, \quad \forall k \neq k_1
\end{equation}
On the other hand, if \(k = k_1\), the expression simplifies to
\begin{align*}
  f_{g,h}^{(k_1)} - P_h^{(k_1)}R_g^{(k_1)} &= \lambda\sqrt{d_g}\frac{f_{g,h}^{(k_1)}}{\norm{\mathbf{f}_{g}^{(k_1)}}} \\
  f_{g,h}^{(k_1)} \left[1 + \frac{\lambda\sqrt{d_g}}{\norm{\mathbf{f}_{g}^{(k_1)}}}\right] &= P_h^{(k_1)}R_g^{(k_1)} \\
  f_{g,h}^{(k_1)} &= \left[1 + \frac{\lambda\sqrt{d_g}}{\norm{\mathbf{f}_{g}^{(k_1)}}}\right]^{-1} P_h^{(k_1)}R_g^{(k_1)}.
\end{align*}
Taking the \(L_2\)-norm on both sides yields
\begin{align*}
  \sqrt{\sum_{h \in [d_g]} \mathbb{E}[(f_{g,h}^{(k_1)})^2]} &= \norm{\mathbf{f}_{g}^{(k_1)}} \\
  &= \left[1 + \frac{\lambda\sqrt{d_g}}{\norm{\mathbf{f}_{g}^{(k_1)}}}\right]^{-1} \norm{\mathbf{Q}R_g^{(k_1)}}.
\end{align*}
Solving for \(\norm{\mathbf{f}_{g}^{(k_1)}}\) gives us the following identity
\begin{equation*}
  \norm{\mathbf{f}_{g}^{(k_1)}} = \norm{\mathbf{Q}R_g^{(k_1)}} - \lambda \sqrt{d_g}.
\end{equation*}
Plugging into the simplified update for above finally yields
\begin{align*}
  f_{g,h}^{(k_1)} &= \left[1 + \frac{\lambda\sqrt{d_g}}{\norm{\mathbf{f}_{g}^{(k_1)}}}\right]^{-1} P_h^{(k_1)}R_g^{(k_1)} \\
  &= \left[1 + \frac{\lambda\sqrt{d_g}}{\norm{\mathbf{Q}R_g^{(k_1)}} - \lambda \sqrt{d_g}}\right]^{-1} P_h^{(k_1)}R_g^{(k_1)} \\
  &= \left[1 - \frac{\lambda\sqrt{d_g}}{\norm{\mathbf{Q}R_g^{(k_1)}}}\right] P_h^{(k_1)}R_g^{(k_1)} \\
  &= \left[\norm{\mathbf{Q}R_g^{(k_1)}} - \lambda\sqrt{d_g}\right] \frac{P_h^{(k_1)}R_g^{(k_1)}}{\norm{\mathbf{Q}R_g^{(k_1)}}}
\end{align*}
for all \(h \in [d_g]\).

In the case where \(m > 1\) entries \(\norm{\mathbf{f}_{g}^{(k_1)}}, \ldots, \norm{\mathbf{f}_{g}^{(k_m)}}\) achieve the sup-norm also simplifies, i.e., for all \(i \in [m]\) we have
\begin{equation*}
  {P}_h^{(k_i)}R_g^{(k_i)} = \lambda\sqrt{d_g}a_i\frac{{f}_{g,h}^{(k_i)}}{\norm{\mathbf{f}_{g}^{(k_i)}}} + {f}_{g,h}^{(k_i)}.
\end{equation*}
Recall that \(\norm{\mathbf{f}_{g}^{(k_1)}} = \cdots = \norm{\mathbf{f}_{g}^{(k_m)}}\) and taking the \(L_2\) norm on both sides as well as summing over all \(m\) yields
\begin{align*}
  \sum_{i=1}^{m}\norm{\mathbf{Q}R_g^{(k_i)}} &= \sum_{i=1}^{m} \norm{\left[\frac{\lambda\sqrt{d_g}a_i}{\norm{\mathbf{f}_{g}^{(k_m)}}} + 1\right] {f}_{g,h}^{(k_i)}} \\
  &= \sum_{i=1}^{m} \left[\frac{\lambda\sqrt{d_g}a_i}{\norm{\mathbf{f}_{g}^{(k_m)}}} + 1 \right] \norm{\mathbf{f}_{g}^{(k_m)}} \\
  &= \left[\frac{\lambda\sqrt{d_g}}{\norm{\mathbf{f}_{g}^{(k_m)}}} \sum_{i=1}^{m}a_i + m \right] \norm{\mathbf{f}_{g}^{(k_m)}} \\
  &= \lambda\sqrt{d_g} + m \norm{\mathbf{f}_{g}^{(k_m)}}.
\end{align*}
Isolating \(\norm{\mathbf{f}_{g}^{(k_m)}}\) gives us the following identity
\begin{equation*}
  \norm{\mathbf{f}_{g}^{(k_m)}} = \frac{1}{m}\left[\sum_{i=1}^{m}\norm{\mathbf{Q}R_g^{(k_i)}} - \lambda\sqrt{d_g}\right] \quad \forall m \in [m].
\end{equation*}
Plugging this into the simplified update for case (3) yields
\begin{multline*}
  f_{g,h}^{(k_m)} = \left[1 + \frac{\lambda\sqrt{d_g}}{\norm{\mathbf{f}_{g}^{(k_m)}}}\right]^{-1} P_hR_g^{(k_m)} = \\
  \Bigg[1 + \frac{\lambda\sqrt{d_g}}{\frac{1}{m}\left[\sum_{i=1}^{m}\norm{\mathbf{Q}R_g^{(k_i)}} - \lambda\sqrt{d_g}\right]}\Bigg]^{-1} P_hR_g^{(k_m)}.
\end{multline*}
After a few algebraic manipulations, we obtain
\begin{equation*}
  f_{g,h}^{(k_m)} = \frac{1}{m}\left[\sum_{i=1}^{m}s_g^{(k_i)} - \lambda\sqrt{d_g}\right] \frac{P_hR_g^{(k_m)}}{s_g^{(k_m)}},
\end{equation*}
where \(s_g^{(k_m)} = \norm{\mathbf{Q}R_g^{(k_m)}}\) for all \(m \in [m]\).

Now that we have the update for case (3), we still need to find the exact condition that the sup-norm is attained. This condition is given in Lemma~\ref{lemma:m_selection}. The arguments follow a similar logic to \citet{Fornasier2008}.

\begin{lemma}\label{lemma:m_selection}
  For some \(m > 1\) precisely \(m\) entries \(\norm{\mathbf{f}_{g}^{(k_1)}}, \ldots, \norm{\mathbf{f}_{g}^{(k_m)}}\) attain the sup-norm \(\max_{k=1, \ldots, {d_j}} \norm{\mathbf{f}_{g}^{(k)}}\) if and only if \[s_g^{(k_m)} \geq \frac{1}{m-1}\left[\sum_{i=1}^{m-1}s_g^{(k_i)} - \lambda\sqrt{d_g}\right]\] and \[s_g^{(k_{m+1})} < \frac{1}{m}\left[\sum_{i=1}^{m}s_g^{(k_i)} - \lambda\sqrt{d_g}\right].\]
\end{lemma}
\begin{proof}
  Assume exactly \(m\) entries \(\norm{\mathbf{f}_{g}^{(k_1)}}, \ldots, \norm{\mathbf{f}_{g}^{(k_m)}}\) attain the sup-norm. Then, for all \(m \in [m]\) we have
  \begin{equation*}
    f_{g,h}^{(k_m)} = \frac{1}{m}\left[\sum_{i=1}^{m}s_g^{(k_i)} - \lambda\sqrt{d_g}\right] \frac{P_hR_g^{(k_m)}}{s_g^{(k_m)}}.
  \end{equation*}
  By Lemmas~\ref{lemma:stationary_cond} and~\ref{lemma:sup_norm} we know that
  \begin{equation*}
    {P}_h^{(k_i)}R_g^{(k_i)} = \lambda\sqrt{d_g}a_i\frac{{f}_{g,h}^{(k_i)}}{\norm{\mathbf{f}_{g}^{(k_i)}}} + {f}_{g,h}^{(k_i)} \quad \forall i \in [m].
  \end{equation*}
  Isolating \(a_i\) after taking the \(L_2\) norm on both sides we obtain
  \begin{equation*}
    a_i = \frac{\norm{\mathbf{Q}R_g^{(k_i)}} - \norm{\mathbf{f}_{g}^{(k_i)}}}{\lambda\sqrt{d_g}} \quad \forall i \in [m].
  \end{equation*}
  Plugging the identity for \(\norm{\mathbf{f}_{g}^{(k_m)}}\) into \(a_m\) and since \(a_m \geq 0\) we obtain
  \begin{equation*}
    \norm{\mathbf{Q}R_g^{(k_m)}} \geq \frac{1}{m-1}\left[\sum_{i=1}^{m-1}\norm{\mathbf{Q}R_g^{(k_i)}} - \lambda\sqrt{d_g}\right].
  \end{equation*}
  Since \(\norm{\mathbf{f}_{g}^{(k_m)}} > \norm{\mathbf{f}_{g}^{(k_{m+1})}}\) we must have that from Eq.~\eqref{eq:case_2_not_k_1} \(\norm{\mathbf{Q}R_g^{(k_{m+1})}} = \norm{\mathbf{f}_{g}^{(k_{m+1})}} < \norm{\mathbf{f}_{g}^{(k_m)}} = \frac{1}{m}\left[\sum_{i=1}^{m}\norm{\mathbf{Q}R_g^{(k_i)}} - \lambda\sqrt{d_g}\right]\).

  On the other hand, assuming that exactly \(l \neq m\) entries \(\norm{\mathbf{f}_{g}^{(k_1)}}, \ldots, \norm{\mathbf{f}_{g}^{(k_l)}}\) attain the sup-norm, then following the same steps as above leads to contradiction.
\end{proof}

From Lemma~\ref{lemma:m_selection} we conclude that there exist exactly \(m^*\) entries that attain the sup-norm if and only if
\begin{equation*}
  m^* = \argmax_m \frac{1}{m} \left[\sum_{i=1}^{m}\norm{\mathbf{Q}R_g^{(k_i)}} - \lambda\sqrt{d_g}\right].
\end{equation*}
This concludes the proof of Theorem~\ref{thm:backfitting_update}.

\section{Simulation Details}\label{sec:simulation_details}

We have developed a Python library that implements all the methodologies presented in this work, including our simulation study. This library is available under the following link \url{https://github.com/boschresearch/gresit}.

\subsection{Synthetic data generation}\label{sec:synthetic_data}

Synthetic data used throughout the experiments is generated from GANMs (Definition~\ref{def:ANM}) with varying dimension and group size. First, we construct the ground truth causal graph using the Erd\"{o}s-R\'{e}nyi model~\citep{Erdos2011} with sparsity level set proportional to the number of nodes. In order to set up the nonlinear link functions \({f}_g\), we follow a strategy similar to previous work on scalar ANMs~\citep{Lachapelle2020, Uemura2022, Rolland2022}. We sample \({f}_g\) from randomly weighted sums of Gaussian processes.

Additive noise is generated from multivariate Log-Normal distributions. First, multivariate Gaussians are generated where mean vector entries and off-diagonal entries in the covariance matrix are sampled uniformly from the \([-0.8,0.8]\) interval. Then the exponential is taken component-wise. While propagating through the graph, we adjust the scales of the variables in each equation to a signal-to-noise ratio of two. To reduce the risk of involuntary design patterns that might be picked up by the algorithms we employ~\citep{Reisach2021}, data is always standardized.

\subsection{Metrics}\label{sec:metrics}

Evaluating causal graphs is a difficult task as the mere number of dissimilar edges between learned and the ground truth graph does not necessarily reflect how the graphs differ in their capability to answer causal queries. This fact is exacerbated by the problem that not all causal discovery algorithms return the same graphical object. Different causal assumptions posed by the corresponding routines lead to different types of causal graphs. Consequently, we report a number of evaluation metrics in order to capture strengths and weaknesses of the routines employed across a wide spectrum of tasks. Next to metrics that quantify edge recovery, we report recently proposed~\citep{Henckel2024} graph distances that count the number of wrongly inferred causal effects, determined by different identification strategies, when using the learned rather than the true DAG.

\textit{Precision, Recall and \(F_1\) score}

We start with binary classification metrics that are prominent in the machine learning literature. Put in a graphical context, \textit{Precision} refers to the fraction of correctly
determined edges among all identified edges. Recall, sets correctly identified edges in relation to all edges present in the
ground truth. The \(F_1\) score is the harmonic mean of Precision and Recall, i.e. \(F_1 =
2 \cdot\text{Precision}\cdot\text{Recall} /(\text{Precision}+\text{Recall})\).

\textit{Structural Hamming Distance (SHD)}

The SHD counts the number of differing edges between to graphs.

\textit{Structural Interventional Distance (SID)}~\citep{Peters2015}:

The SID counts the number of incorrectly inferred interventional distributions from the learned graph when compared to the ground truth. For DAGs, this equates to counting parent sets in the learned DAG that are not valid adjustment sets in the ground truth DAG.

\textit{Ancestor Adjustment Identification Distance (AAID)~\citep{Henckel2024}}

While parent adjustment is a valid adjustment strategy, there exist statistically more efficient adjustment sets. Based on this observation, \citet{Henckel2024} generalize the idea of the SID and present identification distances that count the number of incorrect identification formulas that arise when using some identification strategy. The SID then arises when parent adjustment is chosen as the identification strategy. However, SID may produce surprisingly large quantities when comparing two DAGs that have the same causal order. Choosing ancestor rather than parent adjustment leads to an adjustment strategy that returns a zero whenever estimated and ground truth DAG agree in terms of their causal orders.

\textit{Order Adjustment Identification Distance (OAID)~\citep{Henckel2024}}

In order to emphasize the role played by the pruning step, we also provide a DAG to order distance.~\textit{OAID} arises when comparing the super-DAG \(\mathcal{G}^\pi\) and the ground truth \(\mathcal{G}_0\) in terms of their \textit{AAID}. Consequently,  \textit{OAID} counts the number of incorrect identification formulas derived from the estimated causal order \(\pi\).

\subsection{Algorithms}\label{sec:algorithms}
\begin{table*}[ht]
  \centering
  \caption{Hyperparameter and tuning choices for all methods employed in this work.}
  \label{table:tuning}
  \begin{tabularx}{\textwidth}{@{}lX@{}}
    \toprule
    Method & Parameters \\
    \midrule
    \textit{GroupRESIT} &
    \texttt{regression}: MLP with tanh activation;
    \texttt{n\_epochs}=500;
    \texttt{lr}=0.01;
    \texttt{loss}=MSE;
    \texttt{batch\_size}=500;
    \texttt{indep\_test}=HSIC;
    \texttt{alpha}=0.01
    \\[0.8em]

    \textit{MURGS} &
    \texttt{smoother}=Gaussian kernel regression;
    \texttt{plugin bandwidth}=0.6\,sd$(X)\,n^{-1/5}$
    \\[0.8em]

    \textit{GroupPC} &
    \texttt{indep\_test}=Fisher’s Z;
    \texttt{alpha}=0.05
    \\[0.8em]

    \textit{GroupGraN-DAG} &
    \texttt{hidden\_num}=2;
    \texttt{hidden\_dim}=10;
    \texttt{batch\_size}=64;
    \texttt{lr}=0.001;
    \texttt{iterations}=100\,000;
    \texttt{model\_name}=NonLinGaussANM;
    \texttt{nonlinear}=leaky‐ReLU;
    \texttt{optimizer}=RMSProp;
    \texttt{h\_threshold}=1e-8;
    \texttt{lambda\_init}=0.0;
    \texttt{mu\_init}=0.001;
    \texttt{omega\_lambda}=1e-4;
    \texttt{omega\_mu}=0.9;
    \texttt{stop\_crit\_win}=100;
    \texttt{edge\_clamp\_range}=1e-4
    \\[0.8em]

    \textit{GroupLiNGAM} &
    \texttt{regression}=OLS;
    \texttt{indep\_test}=HSIC
    \\
    \bottomrule
  \end{tabularx}
\end{table*}

We employ the following causal discovery algorithms in our experiments.

\textit{GroupRESIT}

Owing to the modular design of GroupRESIT, one may, in principle, combine various pairs of multi-response regression methods and vector independence tests in the first phase. To ensure broad applicability, we employ neural networks—specifically, multilayer perceptrons (MLPs) with hyperbolic tangent activation functions and early stopping—for the multi-response regression. We use the mean squared error (MSE) loss, which yielded better performance than the HSIC loss \citep{Mooij2009,Greenfeld2020}. For the subsequent independence test, we apply the empirical HSIC~\citep{Gretton2005} with Gaussian RBF kernels, using the median heuristic for bandwidth selection.

In the second phase, we compare the performance of MURGS with that of the greedy independence criterion employed in the original RESIT framework \citep{Peters2014}. Any linear smoother can be used to estimate the conditional expectations during the backfitting procedure; in our experiments, we employ Gaussian kernel regression with a plug-in bandwidth \(h = 0.6\cdot\text{sd}(X)n^{-1/5}\). For greedy independence testing, we again apply the empirical HSIC, with \(p\)-values computed via the gamma approximation on a separate test dataset.

\textit{GroupGraN-DAG}

GraN-DAG~\citep{Lachapelle2020} is a score based algorithm developed to handle nonlinear relations among variables. GraN-DAG utilizes the continuous acyclicity constraint first suggested by~\citet{Zheng2018}. With the appropriate loss function, GraN-DAG can be tailored towards Gaussian nonlinear additive noise models. However, in order to ensure that GraN-DAG operates on a group level, we adapt the micro-level acyclicity constraint to encourage acyclicity on the corresponding group DAG.%, similar to \citet{Kikuchi2023}.

Recall that for variable groups \(\mathbf{X} = (\mathbf{X}_1, \ldots, \mathbf{X}_p)\) each group
\(\mathbf{X}_g \in \mathbb{R}^{d_g}\). Suppose we consider all group entries as scalar random variables such that we have \(m = \sum_{g=1}^p d_g\) many micro variables. GraN-DAG enforces acyclicity via the trace exponential of a weighted adjacency matrix \(A_\phi \in \mathbb{R}_{\geq 0}^{m \times m}\) that arises from the weights in the neural network. More specifically, the micro-acyclicity constraint amounts to \(h(A_\phi) = tr(e^{A_\phi \circ A_\phi}) - m = 0\), where \(\circ\) denotes the Hadamard product. Similar to \citet{Kikuchi2023}, we enforce acyclicity in the corresponding group DAG by the following weighted group adjacency matrix \(A_\phi^{\text{group}} \in \mathbb{R}_{\geq 0}^{p \times p}\) where
\begin{equation}
  (A_\phi^{\text{group}})_{gh} =
  \begin{cases}
    0 &\text{if } g=h\\ \frac{1}{d_g d_h}
    \sum_{i\in [d_g]}\sum_{j\in [d_h]} (A_\phi)_{i,j} &\text{o.w.}
  \end{cases}.
\end{equation}
By setting the diagonal to zero in \(A_\phi^{\text{group}}\), we ignore the graph structure induced
by \(A_\phi\) within the groups and only focus on the inter-group relations.
As we want to enforce an acyclic group graph, we use the following modified constraint in the
augmented Lagrangian method used in~\cite{Lachapelle2020}
\begin{equation}
  h(A_\phi^{\text{group}}) = tr(e^{A_\phi^{\text{group}} \circ A_\phi^{\text{group}}}) - p = 0.
\end{equation}
In general, the final weighted adjacency matrix in~\textit{GroupGraN-DAG} is not sparse. Therefore, appropriate thresholds need to be set to enforce strict zeros. Unfortunately, this might lead to a clipped adjacency matrix that need not necessarily encode a DAG. In such cases, we continue to select thresholds until the resulting weighted adjacency matrix becomes acyclic.

\textit{GroupPC}

The PC algorithm~\citep{Spirtes1993} performs conditional independence tests in a resource efficient way in order to remove edges from a fully connected undirected graph. Given the first phase, the algorithm orients as many edges as possible. We implement the stable version of the algorithm proposed by~\citet{Colombo2014}. In order to adapt the PC algorithm to the group setting, the involved tests need to be able to handle testing conditional independence among two random vectors given a set of random vectors. While~\citet{Zhang2009} extended the HSIC to conditional independence testing its prohibitively long runtime prevents us from using it in our experiments. Instead, we use the simple Fisher-Z scoring test and treat group entries individually. Then, we aggregate the coordinate-wise hypotheses based on scalar variables by using the union-intersection method. More precisely, in the skeleton-finding phase, we remove an edge if the union of the \(p\)-values of the individual tests is larger than the significance level \(\alpha\). Otherwise, the edge is retained. The significance level \(\alpha\) of the involved test acts as a hyperparameter. The smaller \(\alpha\) the larger the hurdle to keep an edge in the first phase such that sparser graphs will be returned. In general, the algorithm returns a completed partially directed acyclic graph (CPDAG). While in principle the new graph metrics developed by \citet{Henckel2024} return meaningful results for CPDAG to DAG comparisons, the same cannot be said for the remaining ones. In particular, comparability becomes difficult between those algorithms that return DAGs and the PC algorithm. Thus, we compute the SHD for each DAG consistent with the CPDAG and select the one with the smallest SHD.

\textit{GroupLiNGAM}

We implement the \textit{GroupDirectLiNGAM} algorithm of~\citet{Entner2012}, which extends the direct estimation method for LiNGAM introduced by~\citet{Shimizu2006,Shimizu2011} to handle vector-valued variables. Since~\citet{Entner2012} focus solely on the causal ordering step, we use MURGS to recover the full graph structure thereafter.

Table~\ref{table:tuning} reports all hyperparameters and tuning parameter choices for benchmark and real data results.

\begin{figure*}[ht!]
  \centering
  \includegraphics[width=.82\textwidth]{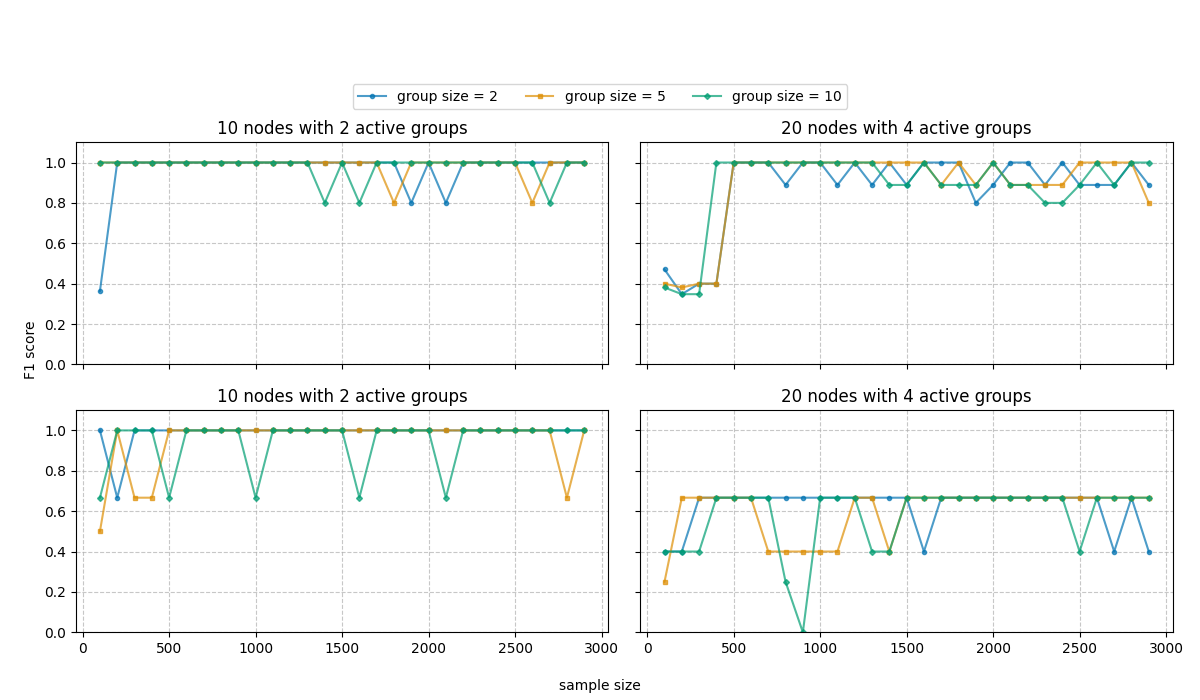}
  \caption{Feature selection capability of MURGS. Node size is \(p=10,20\), with \(2,4\) active groups, respectively.}
  \label{fig:murgs_sim}
\end{figure*}

\section{Additional Results}\label{sec:additional_results}
\paragraph{MURGS experiments}

In this section we report results of a separate simulation study to assess the feature selection capability of MURGS. Figure~\ref{fig:murgs_sim} demonstrates in terms of F1 score (see Section~\ref{sec:metrics}) how MURGS is able to select active groups across with increasing sample size. Encouragingly, group size does not seem to play a significant role in the performance of MURGS. The first row of Figure~\ref{fig:murgs_sim} is based on data generated according to the synthetic data generation strategy described in Sections~\ref{sec:experiments} and~\ref{sec:synthetic_data}. We select sink nodes in a graph making sure that the number of parents is the same across group size. Then we generate data with increasing sample size and track the F1 score. Besides some numerical instabilities MURGS shows convergence behavior already on small sample sizes.

To test out the boundaries, we reiterate the experiments with a different data generation strategy (second row). Nonlinear functions are now generated via randomly initialized deep neural networks with ReLU activation function such that additive models will have a hard time approximating the regression functions. Indeed, MURGS struggles much more to recover the active nodes when the number of nodes in the graph is larger. Despite the challenging nonlinearity F1 score depicts high values throughout again without requiring extraordinarily high sample size.

\paragraph{Additional real data and synthetic data results}

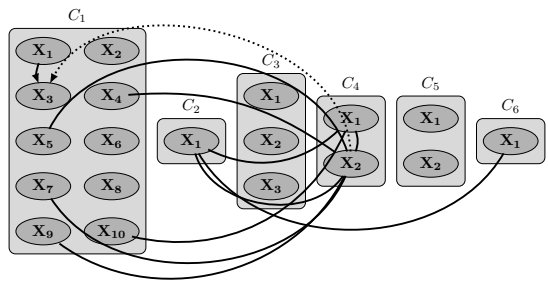
\begin{figure}
  \centering
  \scalebox{0.6}{
    \tikzset{
  group/.append style={
    align=center,
    text=black,
    scale=1.0
  },
  cell/.append style={
    draw, fill=gray!30!white, rounded corners
  }
}

\begin{tikzpicture}

  %C1
  \node [cell, minimum width=3cm, minimum height=5cm, label=$C_1$] at (0,0) {};

  \draw[fill=gray!60!white] (-0.75, 2) ellipse (0.6cm and 0.3cm) coordinate node [group] (C1_X1) {$\mathbf{X_1}$};
  \draw[fill=gray!60!white] (+0.75, 2) ellipse (0.6cm and 0.3cm) coordinate node [group] (C1_X2) {$\mathbf{X_2}$};
  \draw[fill=gray!60!white] (-0.75, 1) ellipse (0.6cm and 0.3cm) coordinate node [group] (C1_X3) {$\mathbf{X_3}$};
  \draw[fill=gray!60!white] (+0.75, 1) ellipse (0.6cm and 0.3cm) coordinate node [group] (C1_X4) {$\mathbf{X_4}$};
  \draw[fill=gray!60!white] (-0.75, 0) ellipse (0.6cm and 0.3cm) coordinate node [group] (C1_X5) {$\mathbf{X_5}$};
  \draw[fill=gray!60!white] (+0.75, 0) ellipse (0.6cm and 0.3cm) coordinate node [group] (C1_X6) {$\mathbf{X_6}$};
  \draw[fill=gray!60!white] (-0.75, -1) ellipse (0.6cm and 0.3cm) coordinate node [group] (C1_X7) {$\mathbf{X_7}$};
  \draw[fill=gray!60!white] (+0.75, -1) ellipse (0.6cm and 0.3cm) coordinate node [group] (C1_X8) {$\mathbf{X_8}$};
  \draw[fill=gray!60!white] (-0.75, -2) ellipse (0.6cm and 0.3cm) coordinate node [group] (C1_X9) {$\mathbf{X_9}$};
  \draw[fill=gray!60!white] (+0.75, -2) ellipse (0.6cm and 0.3cm) coordinate node [group] (C1_X10) {$\mathbf{X_{10}}$};

  \node [cell, minimum width=1.5cm, minimum height=1cm, label=$C_2$] at (2.5,0) {};
  \draw[fill=gray!60!white] (2.5, 0) ellipse (0.6cm and 0.3cm) coordinate node [group] (C2_X1) {$\mathbf{X_1}$};

  \node [cell, minimum width=1.5cm, minimum height=3cm, label=$C_3$] at (4.25,0) {};
  \draw[fill=gray!60!white] (4.25, 1) ellipse (0.6cm and 0.3cm) coordinate node [group] (C3_X1) {$\mathbf{X_1}$};
  \draw[fill=gray!60!white] (4.25, 0) ellipse (0.6cm and 0.3cm) coordinate node [group] (C3_X2) {$\mathbf{X_2}$};
  \draw[fill=gray!60!white] (4.25, -1) ellipse (0.6cm and 0.3cm) coordinate node [group] (C3_X3) {$\mathbf{X_3}$};

  \node [cell, minimum width=1.5cm, minimum height=2cm, label=$C_4$] at (6,0) {};
  \draw[fill=gray!60!white] (6, 0.5) ellipse (0.6cm and 0.3cm) coordinate node [group] (C4_X1) {$\mathbf{X_1}$};
  \draw[fill=gray!60!white] (6, -0.5) ellipse (0.6cm and 0.3cm) coordinate node [group] (C4_X2) {$\mathbf{X_2}$};

  \node [cell, minimum width=1.5cm, minimum height=2cm, label=$C_5$] at (7.75,0) {};
  \draw[fill=gray!60!white] (7.75, 0.5) ellipse (0.6cm and 0.3cm) coordinate node [group] (C5_X1) {$\mathbf{X_1}$};
  \draw[fill=gray!60!white] (7.75, -0.5) ellipse (0.6cm and 0.3cm) coordinate node [group] (C5_X2) {$\mathbf{X_2}$};

  \node [cell, minimum width=1.5cm, minimum height=1cm, label=$C_6$] at (9.5,0) {};
  \draw[fill=gray!60!white] (9.5, 0) ellipse (0.6cm and 0.3cm) coordinate node [group] (C6_X1) {$\mathbf{X_1}$};

  % Directed
  \draw[-latex, black, very thick] (C1_X1) to[bend right=20] (C1_X3);
  \draw[-latex, black, very thick, dotted] (C4_X2) to[bend right=75] (C1_X3);

  % Undirected
  \draw[-, black, very thick] (C4_X2) to[bend right=20] (C1_X4);
  \draw[-, black, very thick] (C4_X2) to[bend left=50] (C1_X9);
  \draw[-, black, very thick] (C6_X1) to[bend left=60] (C2_X1);
  \draw[-, black, very thick] (C4_X1) to[bend left=35] (C2_X1);
  \draw[-, black, very thick] (C4_X2) to[bend right=20] (C4_X1);
  \draw[-, black, very thick] (C1_X5) to[bend left=68] (C4_X2);
  \draw[-, black, very thick] (C1_X10) to[bend right=40] (C4_X1);
  \draw[-, black, very thick] (C4_X2) to[bend left=62] (C1_X7);
  \draw[-, black, very thick] (C4_X2) to[bend left=65] (C2_X1);

\end{tikzpicture}
  }
  \caption{Learned causal edges from the real-world dataset using \textit{GroupPC}.}
  \label{fig:real_data_gpc}
\end{figure}

Figures~\ref{fig:real_data_grandag},~\ref{fig:real_data_gpc}, and~\ref{fig:real_data_glingam} present supplementary results for the remaining algorithms in the real-data experiment (Section~\ref{sec:real_data}) that were omitted from the main text.

Additionally, the ensuing boxplots provide additional results regarding the synthetic experiment described in Sections~\ref{sec:experiments} and~\ref{sec:synthetic_data}. The results plots are ordered by node size and group size.

\clearpage\newpage
\onecolumn

\begin{figure}
  \centering
  \scalebox{0.6}{
    \tikzset{
  group/.append style={
    align=center,
    text=black,
    scale=1.0
  },
  cell/.append style={
    draw, fill=gray!30!white, rounded corners
  }
}

\begin{tikzpicture}

  %C1
  \node [cell, minimum width=3cm, minimum height=5cm, label=$C_1$] at (0,0) {};

  \draw[fill=gray!60!white] (-0.75, 2) ellipse (0.6cm and 0.3cm) coordinate node [group] (C1_X1) {$\mathbf{X_1}$};
  \draw[fill=gray!60!white] (+0.75, 2) ellipse (0.6cm and 0.3cm) coordinate node [group] (C1_X2) {$\mathbf{X_2}$};
  \draw[fill=gray!60!white] (-0.75, 1) ellipse (0.6cm and 0.3cm) coordinate node [group] (C1_X3) {$\mathbf{X_3}$};
  \draw[fill=gray!60!white] (+0.75, 1) ellipse (0.6cm and 0.3cm) coordinate node [group] (C1_X4) {$\mathbf{X_4}$};
  \draw[fill=gray!60!white] (-0.75, 0) ellipse (0.6cm and 0.3cm) coordinate node [group] (C1_X5) {$\mathbf{X_5}$};
  \draw[fill=gray!60!white] (+0.75, 0) ellipse (0.6cm and 0.3cm) coordinate node [group] (C1_X6) {$\mathbf{X_6}$};
  \draw[fill=gray!60!white] (-0.75, -1) ellipse (0.6cm and 0.3cm) coordinate node [group] (C1_X7) {$\mathbf{X_7}$};
  \draw[fill=gray!60!white] (+0.75, -1) ellipse (0.6cm and 0.3cm) coordinate node [group] (C1_X8) {$\mathbf{X_8}$};
  \draw[fill=gray!60!white] (-0.75, -2) ellipse (0.6cm and 0.3cm) coordinate node [group] (C1_X9) {$\mathbf{X_9}$};
  \draw[fill=gray!60!white] (+0.75, -2) ellipse (0.6cm and 0.3cm) coordinate node [group] (C1_X10) {$\mathbf{X_{10}}$};

  \node [cell, minimum width=1.5cm, minimum height=1cm, label=$C_2$] at (2.5,0) {};
  \draw[fill=gray!60!white] (2.5, 0) ellipse (0.6cm and 0.3cm) coordinate node [group] (C2_X1) {$\mathbf{X_1}$};

  \node [cell, minimum width=1.5cm, minimum height=3cm, label=$C_3$] at (4.25,0) {};
  \draw[fill=gray!60!white] (4.25, 1) ellipse (0.6cm and 0.3cm) coordinate node [group] (C3_X1) {$\mathbf{X_1}$};
  \draw[fill=gray!60!white] (4.25, 0) ellipse (0.6cm and 0.3cm) coordinate node [group] (C3_X2) {$\mathbf{X_2}$};
  \draw[fill=gray!60!white] (4.25, -1) ellipse (0.6cm and 0.3cm) coordinate node [group] (C3_X3) {$\mathbf{X_3}$};

  \node [cell, minimum width=1.5cm, minimum height=2cm, label=$C_4$] at (6,0) {};
  \draw[fill=gray!60!white] (6, 0.5) ellipse (0.6cm and 0.3cm) coordinate node [group] (C4_X1) {$\mathbf{X_1}$};
  \draw[fill=gray!60!white] (6, -0.5) ellipse (0.6cm and 0.3cm) coordinate node [group] (C4_X2) {$\mathbf{X_2}$};

  \node [cell, minimum width=1.5cm, minimum height=2cm, label=$C_5$] at (7.75,0) {};
  \draw[fill=gray!60!white] (7.75, 0.5) ellipse (0.6cm and 0.3cm) coordinate node [group] (C5_X1) {$\mathbf{X_1}$};
  \draw[fill=gray!60!white] (7.75, -0.5) ellipse (0.6cm and 0.3cm) coordinate node [group] (C5_X2) {$\mathbf{X_2}$};

  \node [cell, minimum width=1.5cm, minimum height=1cm, label=$C_6$] at (9.5,0) {};
  \draw[fill=gray!60!white] (9.5, 0) ellipse (0.6cm and 0.3cm) coordinate node [group] (C6_X1) {$\mathbf{X_1}$};

  \draw[-latex, black, very thick] (C3_X1) to[bend left=50] (C3_X3);
  \draw[-latex, black, very thick] (C3_X2) to[bend right=10] (C3_X3);

\end{tikzpicture}
  }
  \caption{Learned causal edges from the real-world dataset using \textit{GroupGraN-DAG}.}
  \label{fig:real_data_grandag}
\end{figure}

\begin{figure}
  \centering
  \scalebox{0.6}{
    \tikzset{
  group/.append style={
    align=center,
    text=black,
    scale=1.0
  },
  cell/.append style={
    draw, fill=gray!30!white, rounded corners
  }
}

\begin{tikzpicture}

  %C1
  \node [cell, minimum width=3cm, minimum height=5cm, label=$C_1$] at (0,0) {};

  \draw[fill=gray!60!white] (-0.75, 2) ellipse (0.6cm and 0.3cm) coordinate node [group] (C1_X1) {$\mathbf{X_1}$};
  \draw[fill=gray!60!white] (+0.75, 2) ellipse (0.6cm and 0.3cm) coordinate node [group] (C1_X2) {$\mathbf{X_2}$};
  \draw[fill=gray!60!white] (-0.75, 1) ellipse (0.6cm and 0.3cm) coordinate node [group] (C1_X3) {$\mathbf{X_3}$};
  \draw[fill=gray!60!white] (+0.75, 1) ellipse (0.6cm and 0.3cm) coordinate node [group] (C1_X4) {$\mathbf{X_4}$};
  \draw[fill=gray!60!white] (-0.75, 0) ellipse (0.6cm and 0.3cm) coordinate node [group] (C1_X5) {$\mathbf{X_5}$};
  \draw[fill=gray!60!white] (+0.75, 0) ellipse (0.6cm and 0.3cm) coordinate node [group] (C1_X6) {$\mathbf{X_6}$};
  \draw[fill=gray!60!white] (-0.75, -1) ellipse (0.6cm and 0.3cm) coordinate node [group] (C1_X7) {$\mathbf{X_7}$};
  \draw[fill=gray!60!white] (+0.75, -1) ellipse (0.6cm and 0.3cm) coordinate node [group] (C1_X8) {$\mathbf{X_8}$};
  \draw[fill=gray!60!white] (-0.75, -2) ellipse (0.6cm and 0.3cm) coordinate node [group] (C1_X9) {$\mathbf{X_9}$};
  \draw[fill=gray!60!white] (+0.75, -2) ellipse (0.6cm and 0.3cm) coordinate node [group] (C1_X10) {$\mathbf{X_{10}}$};

  \node [cell, minimum width=1.5cm, minimum height=1cm, label=$C_2$] at (2.5,0) {};
  \draw[fill=gray!60!white] (2.5, 0) ellipse (0.6cm and 0.3cm) coordinate node [group] (C2_X1) {$\mathbf{X_1}$};

  \node [cell, minimum width=1.5cm, minimum height=3cm, label=$C_3$] at (4.25,0) {};
  \draw[fill=gray!60!white] (4.25, 1) ellipse (0.6cm and 0.3cm) coordinate node [group] (C3_X1) {$\mathbf{X_1}$};
  \draw[fill=gray!60!white] (4.25, 0) ellipse (0.6cm and 0.3cm) coordinate node [group] (C3_X2) {$\mathbf{X_2}$};
  \draw[fill=gray!60!white] (4.25, -1) ellipse (0.6cm and 0.3cm) coordinate node [group] (C3_X3) {$\mathbf{X_3}$};

  \node [cell, minimum width=1.5cm, minimum height=2cm, label=$C_4$] at (6,0) {};
  \draw[fill=gray!60!white] (6, 0.5) ellipse (0.6cm and 0.3cm) coordinate node [group] (C4_X1) {$\mathbf{X_1}$};
  \draw[fill=gray!60!white] (6, -0.5) ellipse (0.6cm and 0.3cm) coordinate node [group] (C4_X2) {$\mathbf{X_2}$};

  \node [cell, minimum width=1.5cm, minimum height=2cm, label=$C_5$] at (7.75,0) {};
  \draw[fill=gray!60!white] (7.75, 0.5) ellipse (0.6cm and 0.3cm) coordinate node [group] (C5_X1) {$\mathbf{X_1}$};
  \draw[fill=gray!60!white] (7.75, -0.5) ellipse (0.6cm and 0.3cm) coordinate node [group] (C5_X2) {$\mathbf{X_2}$};

  \node [cell, minimum width=1.5cm, minimum height=1cm, label=$C_6$] at (9.5,0) {};
  \draw[fill=gray!60!white] (9.5, 0) ellipse (0.6cm and 0.3cm) coordinate node [group] (C6_X1) {$\mathbf{X_1}$};

  \draw[-latex, black, very thick] (C1_X9) to[bend right=20] (C1_X7);
  \draw[-latex, black, very thick] (C1_X9) to[bend left=50] (C1_X1);
  \draw[-latex, black, very thick] (C1_X8) to[bend right=40] (C1_X2);
  \draw[-latex, black, very thick] (C1_X4) to[bend left=10] (C1_X5);
  \draw[-latex, black, very thick] (C1_X8) to[bend left=30] (C1_X4);
  \draw[-latex, black, very thick] (C1_X10) to[bend left=50] (C1_X6);

  \draw[-latex, black, very thick] (C3_X1) to[bend left=50] (C3_X3);
  \draw[-latex, black, very thick] (C3_X1) to[bend right=10] (C3_X2);

  \draw[-latex, black, very thick, dotted] (C4_X1) to[bend right=60] (C2_X1);

  \draw[-latex, black, very thick, dotted] (C5_X1) to[bend left=10] (C4_X1);

  \draw[-latex, black, very thick, dotted] (C6_X1) to[bend left=75] (C2_X1);

\end{tikzpicture}
  }
  \caption{Learned causal edges from the real-world dataset using \textit{GroupDirectLiNGAM}.}
  \label{fig:real_data_glingam}
\end{figure}

\begin{centering}
  \includegraphics[width=.82\textwidth]{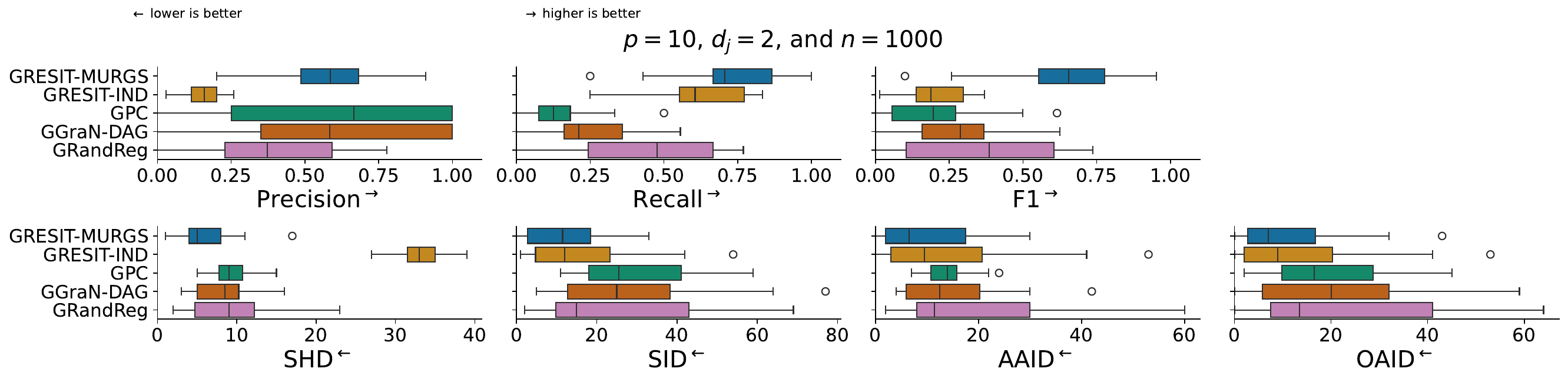}
  \includegraphics[width=.82\textwidth]{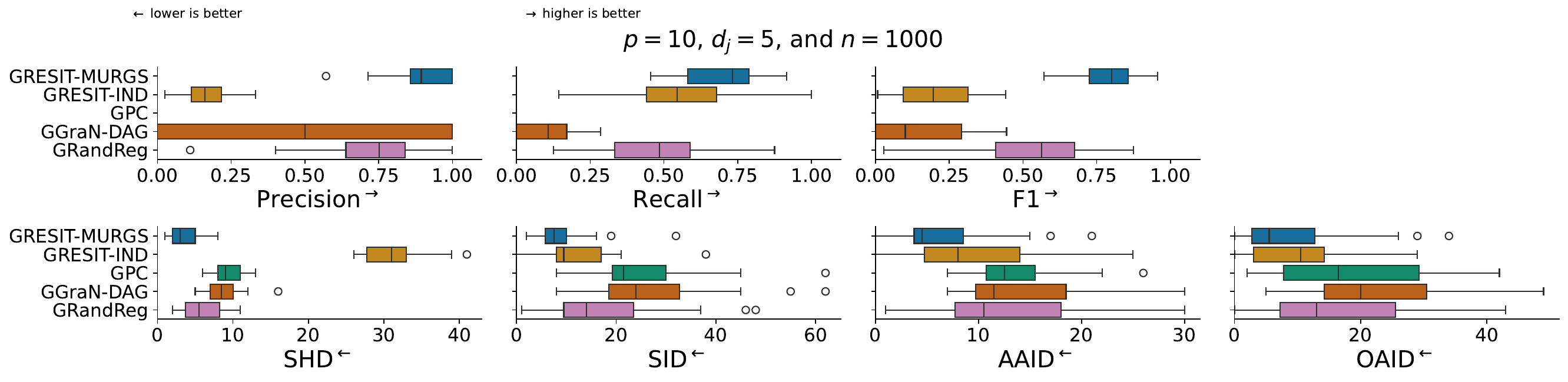}
  \includegraphics[width=.82\textwidth]{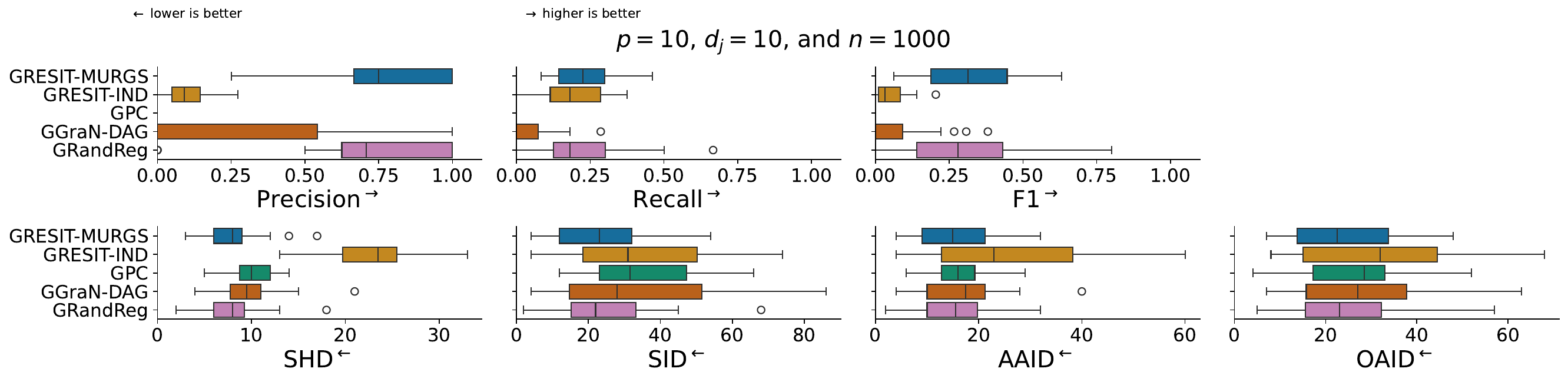}

  \includegraphics[width=.82\textwidth]{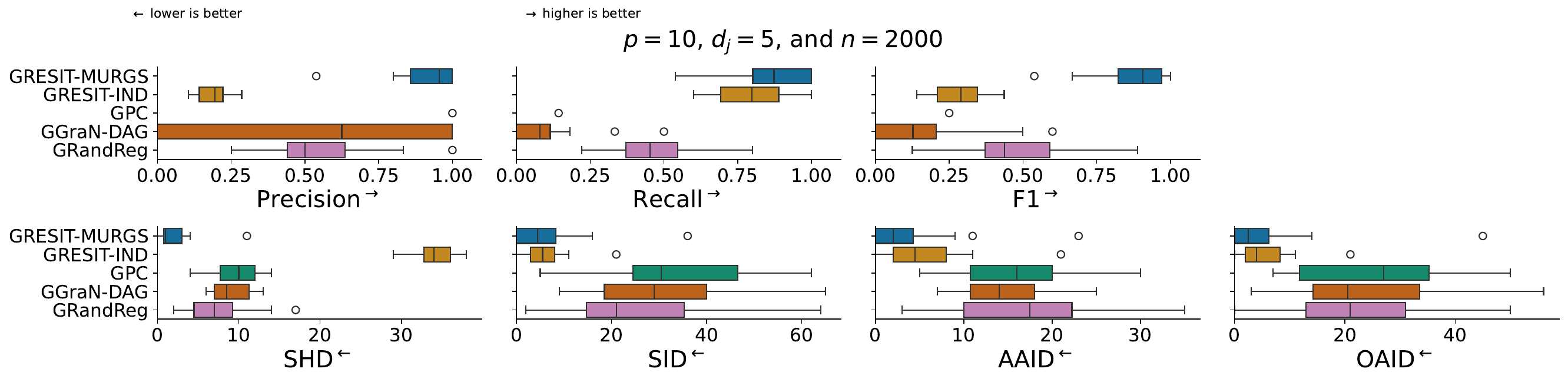}
  \includegraphics[width=.82\textwidth]{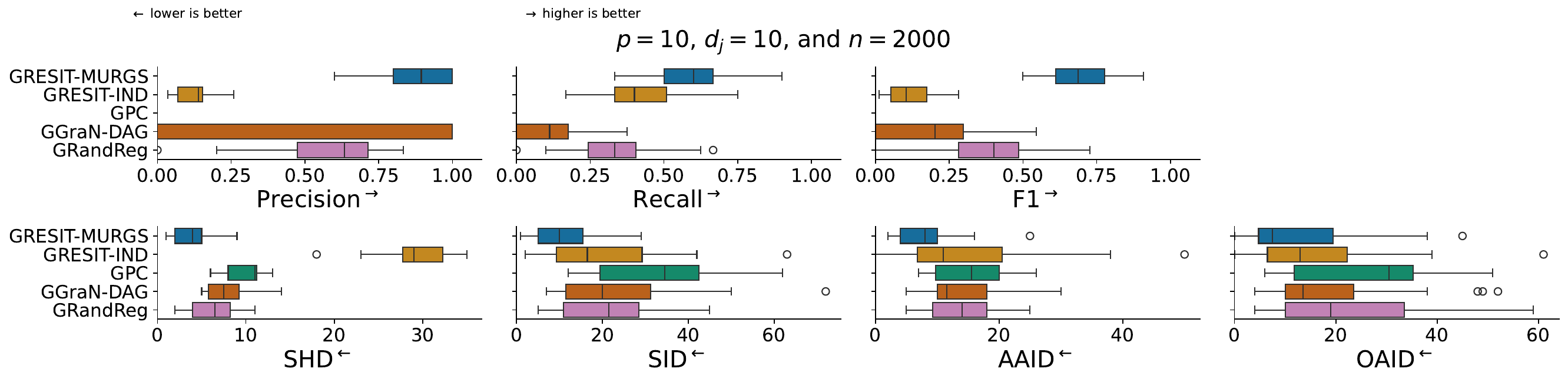}

  \includegraphics[width=.82\textwidth]{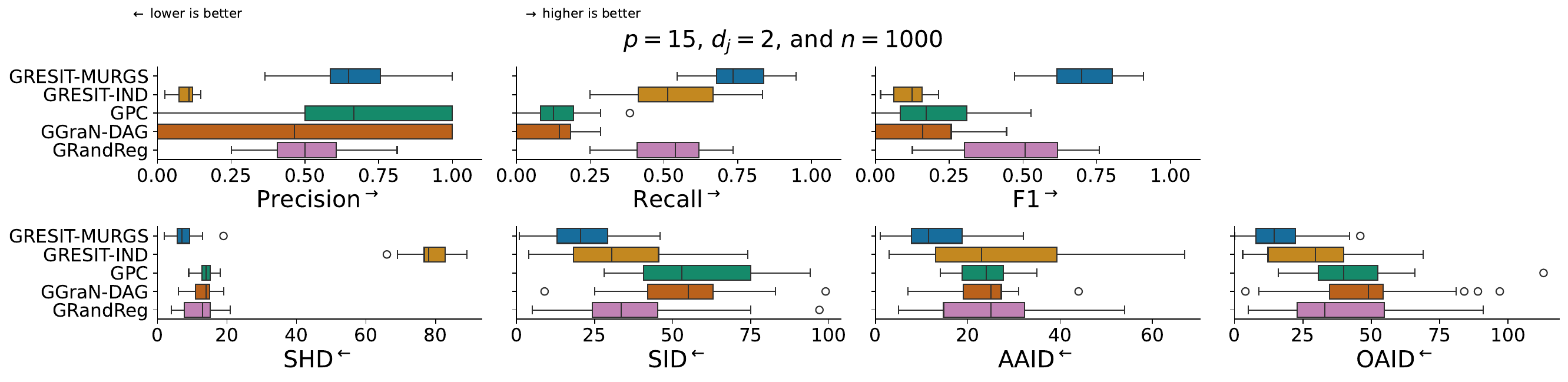}
  \includegraphics[width=.82\textwidth]{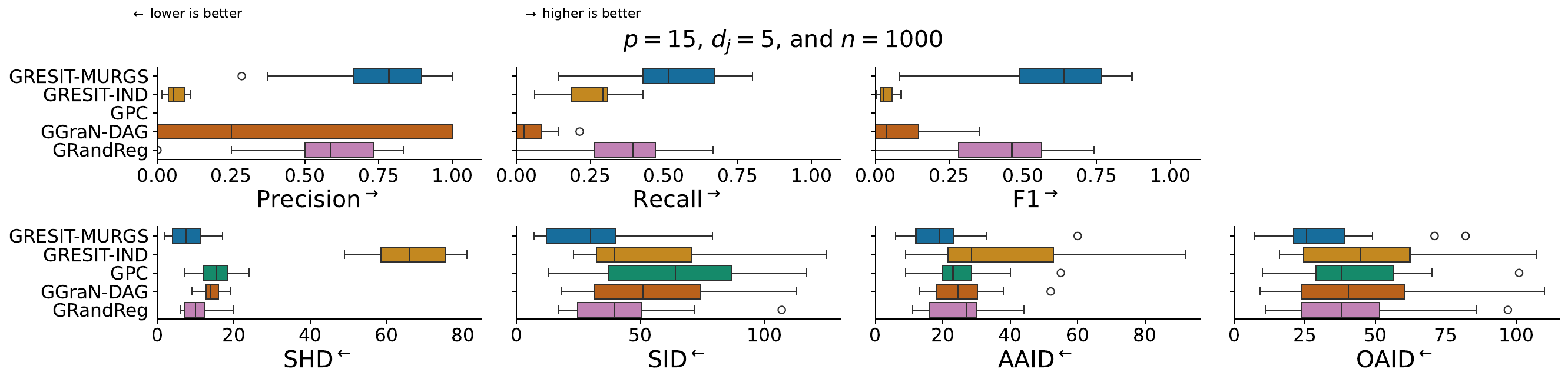}
  \includegraphics[width=.82\textwidth]{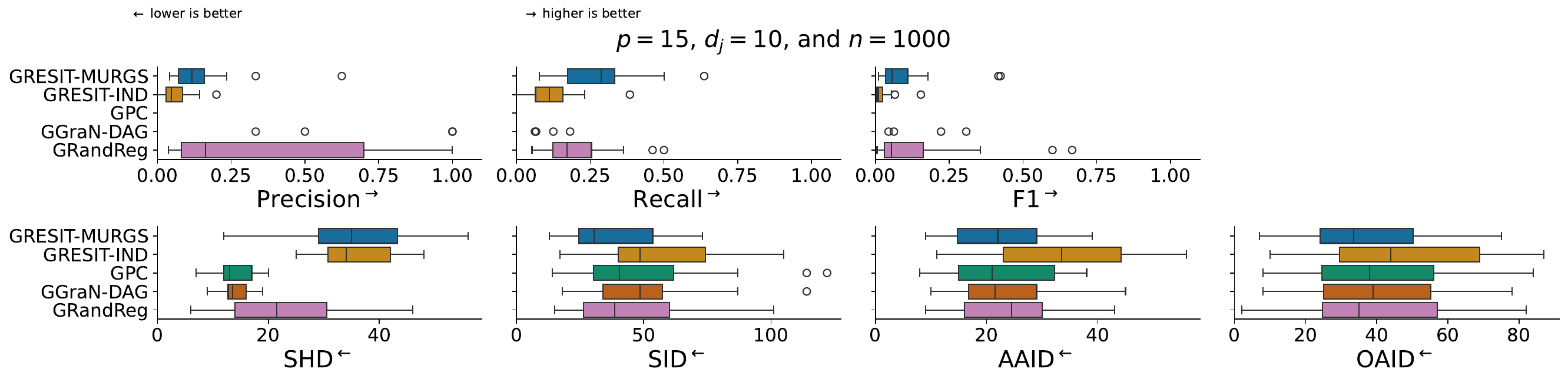}

  \includegraphics[width=.82\textwidth]{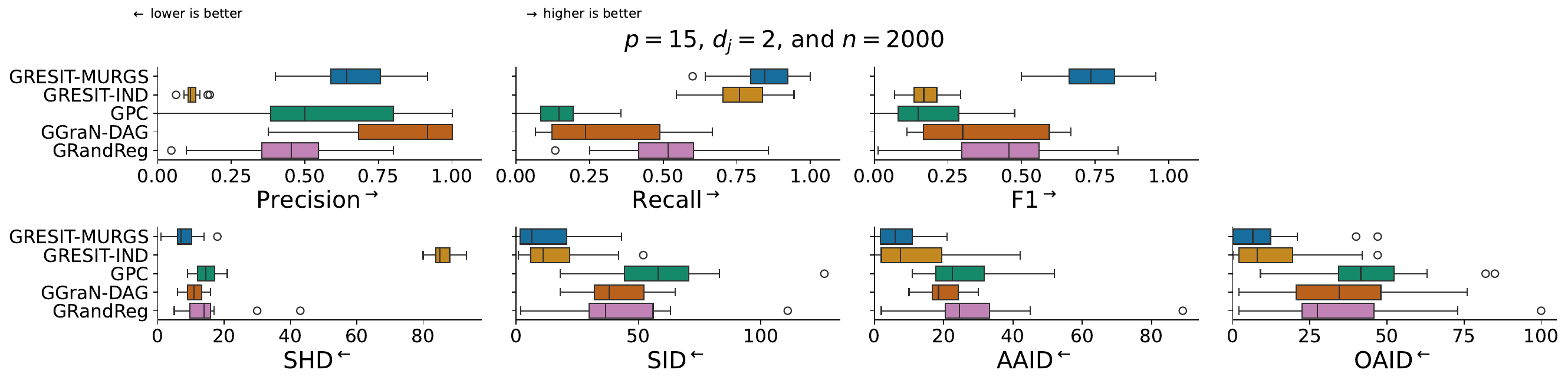}

  \includegraphics[width=.82\textwidth]{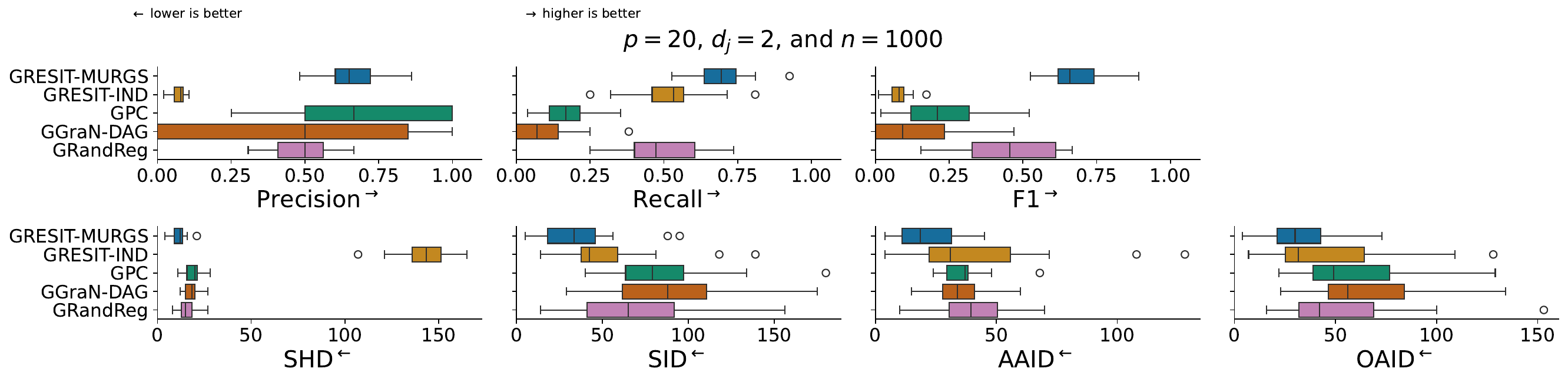}
  \includegraphics[width=.82\textwidth]{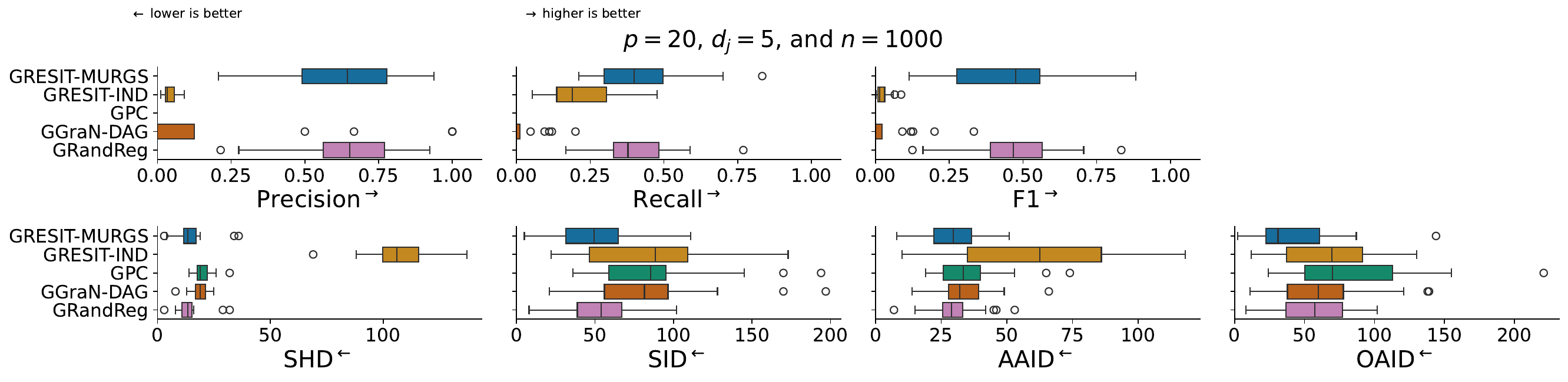}

  \includegraphics[width=.82\textwidth]{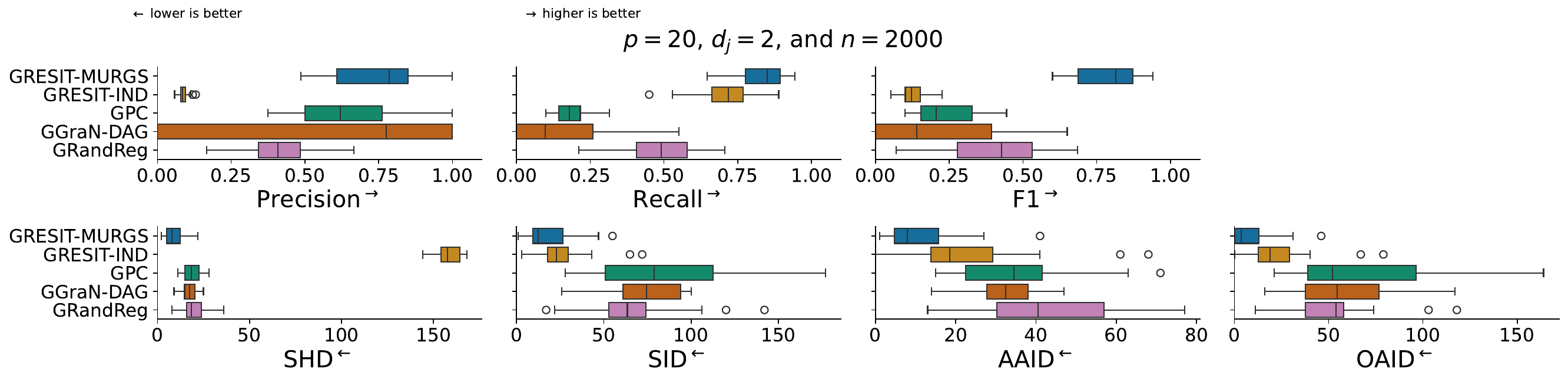}

\end{centering}

\end{document}